\documentclass[11pt]{article}
\usepackage[margin=27mm]{geometry}
\usepackage{amsmath,amssymb,amsthm}
\usepackage{newtxtext,newtxmath}
\usepackage{graphicx,booktabs,array,tabularx,longtable}
\usepackage[authoryear,round]{natbib}
\usepackage{setspace}
\usepackage{microtype}
\usepackage{needspace}
\usepackage{placeins}
\usepackage{xr-hyper}
\usepackage[hidelinks]{hyperref}
\usepackage{url}

\allowdisplaybreaks
\newtheorem{theorem}{Theorem}[section]
\newtheorem{proposition}[theorem]{Proposition}
\newtheorem{lemma}[theorem]{Lemma}
\newtheorem{corollary}[theorem]{Corollary}
\theoremstyle{definition}

\theoremstyle{remark}

\newcommand{\R}{\mathbb R}
\newcommand{\E}{\mathbb E}
\newcommand{\Id}{\operatorname{Id}}
\newcommand{\diag}{\operatorname{diag}}
\newcommand{\tr}{\operatorname{tr}}

\newcommand{\sym}{\operatorname{Sym}}

\graphicspath{{figures/}{../figures/}}

\hypersetup{
pdftitle={Spectral Indistinguishability and Learning-Curve Identifiability in Quadratic Networks},
pdfauthor={Dekun Yang},
pdfkeywords={information-organization sculpting; quadratic neural networks; spectral alignment; learning dynamics; nonlinear observability; signed cycles}}
\title{Spectral Indistinguishability and Learning-Curve Identifiability in Quadratic Networks}
\author{Dekun Yang\\{\normalsize Zhejiang University, China}\\
{\normalsize\texttt{pauliyangwork@gmail.com}}}
\date{}
\begin{document}
\maketitle
\begin{abstract}
Motivated by an information-organization sculpting view of learning,
we study what spectral summaries and scalar learning curves reveal
about centered quadratic networks.
We construct analytic positive input densities with exactly Gaussian
one- and two-coordinate marginals.  Their complete fourth-moment
operators are intertwined by an orthogonal map fixing the entire
diagonal teacher algebra, so the corresponding anchored spectral
measures agree.  For selected common teachers with a nonzero cycle
invariant, their factorized loss curves differ because the map does
not preserve matrix multiplication.
Cycle length $r$, which can be arbitrarily large, gives
$O(\varepsilon^r)$ contrast uniformly in training time for each
fixed construction.
For a supplied pair, noisy scalar loss queries require order
$\varepsilon^{-2r}$ observations, while bounded initialization
interventions with scalar gradient readouts require order
$\varepsilon^{-2}$.
A shared-input score, equal to a paired-gradient difference,
attains the optimal input-sample exponent without knowing the teacher.
An explicit family of diverging auxiliary teacher scales instead
admits a nonzero finite limiting contrast at coupling $1/\log L$.
Conversely, for Gaussian inputs and a rationally independent teacher
spectrum, one exact scalar loss curve identifies the initial predictor
up to sign conjugation.
Within this rationally independent class, with nonzero diagonal
inverse-state error in every coordinate and calibrated time and loss
scales, it also identifies
the unknown teacher spectrum and active dimension, modulo simultaneous
rotations.  For fixed dimension and distinct unordered teacher sums,
dimension-minimal exact observations give generic joint local recovery.
Finite exact global bounds hold for a known teacher and, on the active
class, under a blind dimension cap.
Exact certificates and full-weight replay verify the construction;
numerical inverse checks reveal severe ill-conditioning.
\end{abstract}

\section{Introduction}
We view neural learning as \emph{information-organization sculpting}:
an ongoing process in which learning conditions shape where features
are expressed, how they are distributed and related, and how they
combine to support the network's computations.  Data and tasks,
architecture, initialization, update rules and constraints participate
in this process; the organization already formed influences
subsequent updates.  \mbox{This is an organizing hypothesis about learning.}
It allows strengthening, weakening, differentiation, redistribution
and recombination within a fixed architecture, without assuming that
every form of change occurs in every network.

The mathematical question here concerns one part of this perspective:
whether task-anchored spectral descriptions suffice to determine
nonlinear loss trajectories.  In the quadratic model below, spectral
modes are eigenmatrices of the population moment operator, which acts
on symmetric predictor matrices.  The closed predictor dynamics also
use the Jordan product of the predictor matrix and its loss gradient.
Feature emergence and functional reorganization in ordinary deep
networks remain questions for further work.

Spectral descriptions explain how a learner distributes sensitivity
across modes.  Their explanatory value depends on what information is
retained with those modes.  Eigenvalues alone omit their orientation;
task alignment restores some of that information.  In nonlinear
learning, even a complete family of task-anchored spectral measures
need not specify the Jordan products among the corresponding matrix
responses.

We study this issue in centered quadratic neural networks.  Their
population dynamics close on the positive predictor matrix
$M=WW^\top$, allowing exact comparisons of actual input distributions
and full weight initializations.  Our negative result matches more than
the operator spectrum.  An explicit isometry fixes every diagonal
matrix and intertwines the complete moment operators of two input laws.
Consequently all spectral responses anchored on the diagonal teacher
algebra agree.  The two learners nevertheless have different loss
histories.  The isometry preserves inner products but does not preserve
the multiplication used in the factorized gradient flow.

The converse question gives the second part of the paper.  Under
Gaussian inputs and a known nonresonant teacher, an exact scalar loss
history is a complete invariant of a positive initial predictor, modulo
the unavoidable diagonal sign symmetry.  Its exponential expansion
encodes entry magnitudes and products around hidden cycles.
Within the rationally independent, diagonally active class, the
expansion's observed rate set also determines the unknown teacher
spectrum and active dimension.
For fixed dimension and distinct unordered teacher sums, finitely
many exact scalar observations give generic joint local recovery.
Larger sufficient counts give exact global identification for a known
teacher or for the active blind class with a dimension cap.
These are different observation problems: initial spectral information
need not predict a nonlinear trajectory, whereas the trajectory itself
can reveal information absent from those summaries.

We establish four linked statements.  First, complete spectral
equivalence can fix the whole teacher algebra and still fail to
transport the learning vector field
(Theorem~\ref{thm:geo-equivalence} and
Proposition~\ref{prop:geo-jordan}).  Second, under the nonzero cycle
certificate, the length of a hidden cycle controls the first nonzero
perturbative loss response throughout the entire trajectory for each
fixed construction, with an exact polynomial counterpart
for finitely many weight updates
(Theorems~\ref{thm:geo-response} and \ref{thm:geo-uniform},
Corollary~\ref{cor:geo-discrete-response}).
A bounded initialization intervention reduces the sharp noisy scalar
query exponent from $2r$ to $2$ for the same supplied pair
(Theorem~\ref{thm:geo-channels}); paired gradients give a finite-input
implementation without the teacher
(Corollary~\ref{cor:geo-finite-input}).  Third, an explicitly
unbounded teacher family has a logarithmic amplification threshold:
vanishing moment perturbations create a finite effective state during
a vanishing initial layer (Theorem~\ref{thm:geo-amplification}).
Finally, the scalar Gaussian loss curve identifies that state and,
under an explicit excitation condition, the unknown teacher itself,
while a sharp two-point bound limits
what noisy loss measurements can resolve
(Theorems~\ref{thm:inverse-global}--\ref{thm:inverse-noise}).

\paragraph{Relation to existing theory.}
\citet{patil2023sculpting} introduced \emph{Neural Sculpting}, which
combines pruning during training with network analysis to uncover
hierarchical modular task structure.  The information-organization
perspective here supplies the motivation; the technical contribution
is the spectral and trajectory comparison in a specified quadratic
learner.

Kernel generalization depends on task-model alignment as well as
eigenvalues \citep{canatar2021}.  Our construction matches the full
diagonal-anchored population spectral measures.  Pointwise raw-input
kernel values and empirical Gram matrices remain different observation
objects.

Predictor closure, solvable mode dynamics and factorization-dependent
learning are established \citep{ref6,ref7,ref23,ref9}.
In particular, Neural Quadratic Forms includes both commuting
reductions and a noncommuting isotropic matrix-Riccati solution
\citep[Theorem 7]{ref9}.  We build on that solution to analyze spectral
equivalence and scalar observability.  Extensive-width Gaussian-sensing
theory \citep{ref33} concerns a different asymptotic regime.
Kernel/rich transitions can depend on initialization \citep{woodworth2020};
our amplification result instead keeps the initial factor fixed and
varies teacher scale and the input moment perturbation.

Signed-graph switching and closed walks are classical \citep{ref59,ref64}.
Recovery from magnitudes and cycle products has a particularly close
precedent in principal-minor and determinantal point process learning
\citep{urschel2017,brunel2018,brunel2024}.  The inverse contribution here is the
proof that a scalar learning-loss curve supplies these invariants,
the recovery of the unknown teacher spectrum from its rate set,
and the observation count for joint generic local recovery.
General nonlinear observability \citep{hermann1977} and matrix
factorization geometry \citep{bhatia2019} supply background.
The testing argument uses standard Gaussian divergence inequalities;
the dynamic ingredient is a cycle-dependent signal bound uniform over
all observation times.
General finite-experiment identifiability and exponential-sum
reconstruction are established \citep{sontag2002,potts2010}.
Our blind recovery argument specifies the equivalence class and
excitation conditions associated with this neural loss.

Gradient-query models also distinguish observation channels in sparse
function learning \citep{joshi2024}.  Our comparison concerns
identification of a supplied input-law pair from its neural responses.
The loss and gradient experiments each return one scalar with its
own fixed Gaussian measurement variance.  This makes the exponent
comparison precise without identifying either channel with iid
training data or assigning them equal computational cost.
Separately, a shared-input gradient difference cancels the label
and realizes an ordinary fourth-moment test with an optimal input-sample
exponent for the supplied alternatives.

The principal-minor problem has quadratic query complexity and
approximate recovery guarantees \citep{brunel2024}.  Our fixed-time
observation count has the same dimensional scale and applies to the
scalar neural-loss map.  Table~\ref{tab:contributions} distinguishes
the established reconstruction tools from the dynamic statements.

\begin{table}[t]
\centering\small
\begin{tabular}{p{0.23\textwidth}p{0.32\textwidth}p{0.35\textwidth}}
\toprule
Ingredient & Established result & Dynamic statement proved here\\
\midrule
Quadratic predictor dynamics & Matrix closure and isotropic Riccati
solutions \citep{ref9} & A fixed initial factor develops different effective
states under matched anchored spectra.\\
Signed-cycle invariants & Switching classes, cycle products and
principal-minor recovery \citep{urschel2017,brunel2024} &
One exact scalar loss trajectory determines those invariants for a
rationally independent teacher; diagonal activity also permits blind
spectrum recovery within the active class.\\
Finite observations & Nonlinear observability and quadratic
principal-minor query counts \citep{sontag2002,brunel2024} &
For fixed dimension and distinct unordered teacher sums, any
dimension-minimal set of exact observations at distinct times gives
generic joint local identification.\\
Testing difficulty & Gaussian divergence bounds and cycle-dependent
DPP bounds \citep{urschel2017} &
A loss-signal bound uniform over adaptive observation times.\\
\bottomrule
\end{tabular}
\caption{The contribution is in the learning dynamics and observation
channel.  The listed algebraic and statistical tools are prior results.}
\label{tab:contributions}
\end{table}

\paragraph{Scope.}
The dynamical results concern exact population dynamics of a specified
quadratic learner and constructed laws.  The finite-input test is
restricted to a supplied pair.  The finite limiting contrast requires
diverging teacher norm and initial loss.  The inverse results use
exact observations and nonresonance.  Blind recovery additionally
requires nonzero diagonal inverse-state error; an unexcited coordinate
can remain invisible.  The minimal observation counts are local,
and the bounds valid for every global observation design are
exponential in dimension.  Practical reconstruction on natural data
and transfer to ordinary deep networks remain outside these results.

\section{Model and an admissible paired construction}
\label{sec:model}
For a mean-zero input $x\in\R^d$ with covariance $I$, define
\[
 A(x)=\frac{xx^\top-I}{\sqrt2},\qquad
 f_W(x)=\langle WW^\top,A(x)\rangle,\qquad
 y(x)=\langle T,A(x)\rangle,\quad T\succ0.
\]
The columns of $W$ are quadratic neural units.  With squared population
loss, the complete moment operator and training equations are
\begin{align}
 H(E)&=\E[A(x)\langle A(x),E\rangle],&
 G(M)&=H(M-T),\nonumber\\
 \dot W&=-2G(WW^\top)W,&
 \dot M&=-2\{G(M),M\},\quad M=WW^\top.
 \label{eq:paper-flow}
\end{align}
We take the same square full factor $W_0=\sqrt cI$ in both arms.
The exact predictor update induced by weight gradient descent is
\[
 M_{k+1}=M_k-2h(G_kM_k+M_kG_k)+4h^2G_kM_kG_k.
\]
Dropping the last term would change the optimizer.

\begin{table}[t]
\centering\small
\begin{tabular}{p{0.28\textwidth}p{0.64\textwidth}}
\toprule
Symbol & Meaning and context\\
\midrule
$A(x)$ & Centered quadratic feature matrix.\\
$A,C$ & Auxiliary and interaction coordinate sets in the paired-law construction.\\
$B_\pm$ & Signed interaction adjacency matrices defining the two input laws.\\
$C=M_0^{-1}-T^{-1}$ & Inverse-state error matrix, used in the Gaussian inverse problem.\\
$A_\pm$ & Cycle matrices defining a supplied initial-state pair in the noisy inverse test.\\
$t,s$ & Training time in the forward construction and Gaussian inverse problem, respectively.\\
$t_i$ & Teacher eigenvalues in the inverse problem.\\
\bottomrule
\end{tabular}
\caption{Notation across the construction and inverse problem.
The coordinate set $C$ and inverse-state matrix $C$ occur in separate
sections.  Arguments and subscripts distinguish $A(x)$ and $A_\pm$
from the coordinate set $A$.}
\label{tab:notation}
\end{table}

Fix $q,r\ge3$ and split $d=q+2r-1$ coordinates into an auxiliary set
$A$ and interaction set $C$.  On $C$, two chordless $r$-cycles share
one hub.  Corresponding edges have the same positive weights
$w_1,\ldots,w_r$.  In $B_+$ the two cycle products have signs $(+,-)$;
in $B_-$ they have signs $(-,+)$.  Choose the changed edge in each
branch correspondingly.  The branch-exchange involution $P$ satisfies
$B_-=PB_+P^\top$, while $|B_-|=|B_+|$.

Let $G_0=\diag(g)$ have distinct negative entries and let $T=cI-G_0$.
For curve separation we additionally choose $g$ so that
\[
 J_r=\tr\!\left[\diag(g_C)(B_+^r-B_-^r)\right]\ne0.
\]
By \eqref{eq:isospectralJ}, this is equivalent to unequal sums of the
teacher errors on the two branches excluding the hub.  Distinct
negative entries alone do not ensure it.  Choosing all errors on one
branch closer to zero than those on the other gives an admissible
noncancelling choice.
The auxiliary barycentric vector
\[
 \ell_i=\prod_{\substack{j\in A\\j\ne i}}(g_i-g_j)^{-1}
\]
satisfies $\ell^\top g_A^k=0$ for $k<q-1$ and
$\ell^\top g_A^{q-1}=1$.  In particular
$\ell^\top\mathbf1=\ell^\top T_A=0$.
For physical rescalings below, only the specified normalization changes;
the moment-operator formula remains explicit.

\paragraph{Positive input laws.}
Let $p_0$ be the product five-node Gaussian quadrature law for a standard
normal input.  In one coordinate its nodes are
$0,\pm\sqrt{5-\sqrt{10}},\pm\sqrt{5+\sqrt{10}}$;
the mass at zero is $8/15$, at each inner node
$(7+2\sqrt{10})/60$, and at each outer node
$(7-2\sqrt{10})/60$.  Its moments through degree nine match the
standard normal moments.  Set
\begin{equation}
 h_s(x)=\frac12\left(\sum_{i\in A}\ell_i x_i^2\right)
             \sum_{e\in E(C)}(B_s)_e x_{e_1}x_{e_2},
 \qquad p_s(x)=p_0(x)(1+\varepsilon h_s(x)).
 \label{eq:paper-law}
\end{equation}
If $|\varepsilon|R_5<1$, where
\[
 R_5=\frac{(5+\sqrt{10})^2}{4}\|\ell\|_1
                        \sum_{e\in E(C)}w_e,
\]
every mass is positive.  The bound uses $\sum\ell_i=0$ and the
largest squared node.  Oddness at interaction endpoints and
$\sum\ell_i=0$ show that the tilted law has mass one, zero mean and
covariance $I$.  Fourth-moment contractions against the tilt have
coordinate degrees at most eight, so Gaussian quadrature is exact.
They give the complete operator in \eqref{eq:geo-complete}, including
the auxiliary--interaction directions.  Thus the operator is realized
by actual input laws, not merely a proposed positive linear map.

\Needspace{6\baselineskip}
\begin{proposition}[Exact realization by analytic Gaussian perturbations]
\label{prop:geo-smooth}
For the fixed construction there are bounded analytic functions $\psi_s$
such that, for all sufficiently small real $\varepsilon$,
\[
 p_{s,\varepsilon}(x)=\phi_d(x)(1+\varepsilon\psi_s(x))
\]
is an everywhere positive density with complete moment operator
$\Id+\varepsilon K_s$.  Every one- or two-coordinate marginal is
exactly the corresponding standard Gaussian marginal.
For an explicit constant $R_\psi<\infty$,
\[
 (1-|\varepsilon|R_\psi)\phi_d\le p_{s,\varepsilon}
       \le(1+|\varepsilon|R_\psi)\phi_d,\qquad
 \|P_{s,\varepsilon}-N(0,I)\|_{\rm TV}
       \le|\varepsilon|R_\psi/2.
\]
In particular, all subsequent exact dynamics and spectral identities
hold under these smooth laws without an approximation of the moments.
\end{proposition}
\begin{proof}
Let $X\sim N(0,1)$ and define the bounded analytic functions
\[
 u(x)=\frac{1-e\cos(\sqrt2 x)}{2},\qquad
 v(x)=\sqrt e\sin x.
\]
The Gaussian characteristic function and its first two derivatives give
\[
 \mathbb E u(X)=0,\quad \mathbb E[Xu(X)]=0,\quad
 \mathbb E[X^2u(X)]=1,\qquad
 \mathbb E v(X)=0,\quad \mathbb E[Xv(X)]=1 .
\]
Set
\[
 \psi_s(x)=
 \left(\sum_{i\in A}\ell_i u(x_i)\right)
 \sum_{(j,k)\in E(C)}(B_s)_{jk}v(x_j)v(x_k).
\]
Each perturbation term is a product of centered functions on three
distinct coordinates.  Integrating out all but any one or two
coordinates therefore removes every term.  This proves the
Gaussian marginal statement and, in particular, zero mean and
covariance $I$.
For a product monomial to have a nonzero expectation against one
term of this perturbation, its auxiliary coordinate must occur
at least twice and both interaction endpoints at least once.
All moments of degree at most three therefore remain Gaussian.
At degree four the only possibilities are
\[
 \int\phi_d(x)\psi_s(x)x_i^2x_jx_k\,dx
   =\ell_i(B_s)_{jk},\qquad i\in A,\ (j,k)\in E(C).
\]
These are exactly the fourth moments generating
\eqref{eq:geo-complete}.  All other lower or fourth moments agree
with the standard Gaussian.

Since $\sum_i\ell_i=0$, the constant part of $u$ cancels in its
weighted sum.  Consequently the explicit bound is
\[
 \|\psi_s\|_\infty\le
 R_\psi:=\frac{e^2}{2}\|\ell\|_1\sum_{e\in E(C)}w_e .
\]
For $|\varepsilon|R_\psi<1$ this proves positivity, normalization,
the density-ratio bounds and the total-variation bound.
\end{proof}

Products of centered bounded functions are an established way to
construct distributions with prescribed marginals, as in the
Sarmanov family and its extensions \citep{lee1996}.
Likewise, pairwise independence without joint independence and
three-variable interaction tests are established
\citep{sejdinovic2013}.
The role of this elementary analytic construction is to realize the
specific complete neural moment operator exactly.
For the logarithmic amplification
family below, these functions and $R_\psi$ are independent of $L$.
Consequently those laws converge even in uniform density ratio to
the actual Gaussian, while the teacher norm diverges.  The finite
contrast is therefore not a consequence of atoms or vanishing
support probabilities.  The construction gives no dimension-uniform
lower bound on the admissible coupling.

\begin{proposition}[The marginal-order boundary for diagonal training]
\label{prop:geo-marginal-boundary}
In a fixed coordinate basis, suppose every marginal on at most three
coordinates of an input law is standard Gaussian.
For any diagonal teacher $T$ and positive diagonal initial predictor,
its factorized population flow and exact weight-gradient-descent
predictor sequence coincide with those under the full standard
Gaussian law, for the same initialization and step sizes.
Both original-law and Gaussian-evaluation losses coincide as well.
Requiring only the one- and two-coordinate marginals to be Gaussian
does not have this implication.
\end{proposition}
\begin{proof}
For every diagonal $D$,
\[
 H(D)_{ij}
 =\frac12\sum_k D_{kk}\,
  \mathbb E[(x_ix_j-\delta_{ij})(x_k^2-1)] .
\]
Each expectation depends on at most three coordinates, so it equals
its Gaussian value.  Thus $H(D)=D$.
The diagonal predictor subspace is invariant under both the flow
and the exact predictor update in Section~\ref{sec:model}; on this
subspace their equations are the Gaussian equations.  Uniqueness
of the flow and induction over updates prove equality.
For diagonal error $E$, the loss is
$\langle E,H(E)\rangle/2=\|E\|_F^2/2$.
For the final assertion, Proposition~\ref{prop:geo-smooth} gives
exactly Gaussian two-coordinate marginals, whereas the nonzero
cycle certificate below gives different learning curves for
admissible diagonal teachers and a common diagonal initialization.
\end{proof}

This boundary concerns the specified diagonal training protocol
and coordinate basis.  It does not state that three-coordinate
marginals determine the complete moment operator or the dynamics
from arbitrary nondiagonal states.

The diagonal auxiliary/interaction block space is invariant.  Writing
$M_s=\diag(a_s)\oplus N_s$,
$S_s=\ell^\top a_s$, $Q_s=\tr(B_sN_s)$ gives
\begin{align}
 \dot a_{s,i}
 &=-4a_{s,i}(a_{s,i}-T_i)-2\varepsilon\ell_i a_{s,i}Q_s,\nonumber\\
 \dot N_s
 &=-4N_s^2+2\{T_C,N_s\}-\varepsilon S_s\{B_s,N_s\}.
 \label{eq:intrinsicblockflow}
\end{align}
The restriction of $K_s=(H_s-\Id)/\varepsilon$ is
\begin{equation}
 K_s(\diag(e_A)\oplus E_C)=
 \tfrac12\diag(\ell\,\tr(B_sE_C))
 \oplus\tfrac12(\ell^\top e_A)B_s.
 \label{eq:intrinsicblockoperator}
\end{equation}
It gives $H_s(I)=I$, $H_s(T)=T$ and $H_s(G_0)=G_0$.
We distinguish original-law training loss and evaluation of both
predictors under one common standard Gaussian law:
\begin{align}
 L_s^G&=\tfrac12\sum_{i\in A}(a_{s,i}-T_i)^2+
                         \tfrac12\|N_s-T_C\|_F^2,\nonumber\\
 L_s^P&=L_s^G+\tfrac{\varepsilon}{2}S_sQ_s.
 \label{eq:intrinsiclosses}
\end{align}
Their initial values agree between arms.

\section{Spectral equivalence with the entire teacher algebra fixed}
\label{sec:geometry}

We use the complete moment operators from the two-cycle construction,
$H_s=\Id+\varepsilon K_s$, $s\in\{+,-\}$.  Write a symmetric matrix as
\[
 E=\begin{pmatrix} E_A&X\\X^\top&Y\end{pmatrix},\qquad
 a=\diag(E_A).
\]
The complete perturbation is
\begin{equation}
 (K_sE)_{A,ii}=\frac12\ell_i\tr(B_sY),\quad
 (K_sE)_C=\frac12(\ell^\top a)B_s,\quad
 (K_sE)_{A,C}=\diag(\ell)XB_s,
 \label{eq:geo-complete}
\end{equation}
with zero remaining auxiliary components.  The two weighted $r$-cycles
share their hub and have opposite sign products.  A permutation $P$
exchanges the branches, so $B_-=PB_+P^\top$.  All teachers in this section
are diagonal in these fixed coordinates.

\begin{theorem}[Complete equivalence of diagonal-anchored spectral data]
\label{thm:geo-equivalence}
There is a Frobenius-orthogonal involution $U$ on $\sym(d)$ such that
\[
 UI=I,\qquad UE=E\quad\text{for every diagonal }E,\qquad
 H_-=UH_+U^*.
\]
Consequently, for any diagonal $E,F$ and any real functions $f,g$ defined
on the finite spectrum,
\begin{equation}
 \langle f(H_+)E,g(H_+)F\rangle
 =\langle f(H_-)E,g(H_-)F\rangle.
 \label{eq:geo-anchored}
\end{equation}
This includes all orders of the operator Krylov response Grams, the
matrix-valued spectral measure anchored on the complete diagonal algebra,
and its resolvents.  It includes the teacher and initial-error spectral
weights of the population initial tangent kernel, whose nonzero
feature-space operator is $4cH_s$ at $W_0=\sqrt cI$.
\end{theorem}
\begin{proof}
For supported interaction edges put
$\sigma_{ij}=(B_-)_{ij}/(B_+)_{ij}\in\{-1,1\}$, and put $\sigma_{ij}=1$
on the diagonal and unsupported entries.  Let
$V(Y)=(\sigma_{ij}Y_{ij})$.  Define
\[
 U\begin{pmatrix}E_A&X\\X^\top&Y\end{pmatrix}
 =\begin{pmatrix}E_A&XP^\top\\PX^\top&V(Y)\end{pmatrix}.
\]
Both components are orthogonal involutions, $VB_+=B_-$, and $U$ fixes
every diagonal matrix.  The first two components of
\eqref{eq:geo-complete} intertwine because
$\langle B_-,VY\rangle=\langle B_+,Y\rangle$.
The cross component intertwines because
$XP^\top B_-=XB_+P^\top$.  Thus $UK_+=K_-U$.
Functional calculus and orthogonality give \eqref{eq:geo-anchored}.
Finally,
$\nabla_W f_W(x)=2A(x)W$ gives the initial kernel
$4c\langle A(x),A(x')\rangle$; its nonzero eigenvalues are those of
$4cH_s$.  The teacher and initial error are diagonal anchors.
\end{proof}

This theorem concerns population spectral information.  It does not say
that the two kernels agree pointwise on raw inputs, that empirical Gram
matrices coincide, or that a coordinate-resolved query $H(E)$ is hidden.
In fact a suitable diagonal query returns the different matrices $B_s$.
Nor does it identify the full initial parameter Hessian spectra.

\begin{proposition}[The multiplication condition for universal transport]
\label{prop:geo-jordan}
Let $U:\sym(d)\to\sym(d)$ be a unital Frobenius isometry, and write
$E\circ F=(EF+FE)/2$.  The following conditions are equivalent:
\begin{enumerate}
\item $U(E\circ F)=UE\circ UF$ for all $E,F$.
\item $U$ transports the factorized vector field
$F_{H,b}(M)=-4M\circ(HM-b)$ to
$F_{UHU^*,Ub}(UM)$ for every self-adjoint $H$, every $b$, and every $M$.
\item Such transport holds for $H=\Id$, $b=\tau I$, and every $M$ in
an open neighborhood of a positive scalar matrix.
\end{enumerate}
Under these conditions it also transports the exact predictor update
$(I-2hG)M(I-2hG)$ of weight gradient descent.
\end{proposition}
\begin{proof}
The first condition directly implies the second; the second implies the
third.  Under the third, cancellation of the linear term in
$F_{\Id,\tau I}(M)=-4M^2+4\tau M$ gives $U(M^2)=(UM)^2$ on an open set.
This polynomial identity holds everywhere.  Polarization proves the first
condition.  The exact update follows from
$GMG=2G\circ(G\circ M)-G^2\circ M$.
\end{proof}

This proposition applies a classical Jordan-preservation condition to the
learning vector field; it is not a new classification of Jordan
automorphisms or Bures isometries.  On real symmetric matrices these
automorphisms are orthogonal conjugations.  In the diagonal-fixed case
one can see the obstruction directly: an automorphism fixing all diagonal
projectors must send $E_{ij}+E_{ji}$ to
$\epsilon_i\epsilon_j(E_{ij}+E_{ji})$, and therefore preserve every named
cycle product.  The $U$ in Theorem~\ref{thm:geo-equivalence} changes these
products and is not such an automorphism.

For this $U$, on the invariant training space
$\mathcal B=\diag(\sym(q))\oplus\sym(2r-1)$, diagonal Jordan
multiplication commutes with $U$.  Thus the affine linearizations about
any common diagonal reference path are intertwined.  Quadratic
multiplication of two nondiagonal state components is the additional
operation they omit.  This assertion is restricted to $\mathcal B$.
On the auxiliary--interaction directions the complete Jacobian contains
the different pencils $D_C+2c\varepsilon\ell_iB_s$; their spectral moments
can distinguish the arms.

\section{Cycle order throughout training and under interventions}
\label{sec:geometry-response}

Fix a teacher, initialization and graph size.  Let
$L_s^G(t,\varepsilon)=\frac12\|M_s(t,\varepsilon)-T\|_F^2$
and
$L_s^P(t,\varepsilon)=\frac12\langle M_s-T,H_s(M_s-T)\rangle$.
The fixed teacher must satisfy $\ell^\top T_A=0$.

\begin{theorem}[Signed-cycle order of the entire weak-coupling response]
\label{thm:geo-response}
For every finite $T_0>0$ there is $\varepsilon_0>0$ for which, uniformly
on $0\le t\le T_0$,
\begin{equation}
 L_+^\zeta(t,\varepsilon)-L_-^\zeta(t,\varepsilon)
   =\varepsilon^r\Phi_r^\zeta(t)+O_{T_0}(\varepsilon^{r+2}),
 \qquad \zeta\in\{G,P\}.
 \label{eq:geo-beta}
\end{equation}
Every lower $\varepsilon$ coefficient vanishes identically in time.
Only powers congruent to $r$ modulo two can occur in the difference.
For the normalized auxiliary barycentric construction with
$J_r=\tr[D_C(B_+^r-B_-^r)]\ne0$,
\[
 \Phi_r^\zeta(t)=\overline C_{q,r}^\zeta t^{rq}+O(t^{rq+1}),
 \qquad \overline C_{q,r}^\zeta=C_{q,r}^\zeta/\varepsilon^r\ne0,
\]
where $C_{q,r}^\zeta$ is the exact initial-time coefficient.
In particular, on a sufficiently short compact interval
$[t_0,t_1]\subset(0,\infty)$, all sufficiently small positive couplings
give a loss difference with a fixed nonzero sign.
\end{theorem}
\begin{proof}
At zero coupling the positive diagonal logistic solution exists on every
finite interval.  Analytic dependence of a polynomial ODE on parameters,
continued over a finite cover of that interval, gives a common complex
parameter neighborhood and uniformly convergent coupling expansions.
This also gives a uniform remainder bound.

Assign an independent variable to each signed edge.  Each coefficient of
degree $k$ in the coupling is a homogeneous polynomial of total degree
$k$ in these edge variables.  Changing signs at interaction vertices
conjugates the state by the associated diagonal sign matrix and leaves
both scalar losses invariant.  Consequently every nonzero edge monomial
has even total incident degree at each vertex.  Its odd-multiplicity
edges form an even subgraph.  On the two-cycle graph the possibilities
are the empty set, either individual cycle, or their union.

The empty set and the union give the same sign for cycle products
$(+,-)$ and $(-,+)$.  A differing monomial must therefore have exactly
one whole cycle as its odd edge set.  Its degree is $r+2m$, $m\ge0$.
This proves the vanishing and parity assertions, including the
$r+2$ remainder.  The initial-time coefficient identifies the first
nonzero Taylor coefficient of $\Phi_r^\zeta$.  Analyticity and compactness
then give the final fixed-sign assertion.
\end{proof}

The fixed teacher is essential to the uniformity in
Theorem~\ref{thm:geo-response}.  The next result determines an explicit
family in which this uniformity fails.

\begin{corollary}[Exact divisibility for a finite number of weight updates]
\label{cor:geo-discrete-response}
For every fixed finite number of actual weight-gradient steps and every
fixed step size, the between-arm difference of either loss is a polynomial
of the form
\[
 L_{+,K}^{\zeta}(\varepsilon)-L_{-,K}^{\zeta}(\varepsilon)
       =\varepsilon^r P_{K,h}^{\zeta}(\varepsilon^2).
\]
The polynomial is permitted to be zero, for example before the hidden
interaction has been reached.
\end{corollary}
\begin{proof}
Each update $W\mapsto W-2hH_s(WW^\top-T)W$ is polynomial in the weights
and edge variables.  A finite number of updates and either quadratic
predictor loss are therefore polynomials.  Diagonal sign conjugation
of the interaction coordinates preserves the initialization, teacher,
update and scalar loss.  The even-incidence and two-cycle parity argument
in the preceding proof applies to every monomial of this polynomial,
and gives the stated divisibility without an asymptotic remainder.
\end{proof}

\subsection{An infinite-horizon bound and the value of an intervention}
\label{sec:geo-alltime}

For a fixed bounded teacher, late observations do not evade the
cycle-order obstruction.  The next theorem strengthens the compact-time
statement to the entire training trajectory.

\begin{theorem}[Uniform cycle order throughout training]
\label{thm:geo-uniform}
Fix the construction, $T\succ0$, $c>0$ and a positive-law realization.
There are $\varepsilon_0,C,\beta>0$ such that, for both losses and all
$|\varepsilon|\le\varepsilon_0$, $t\ge0$,
\[
 L_+^\zeta(t,\varepsilon)-L_-^\zeta(t,\varepsilon)
   =\varepsilon^r\Phi_r^\zeta(t)
                 +\varepsilon^{r+2}R_\zeta(t,\varepsilon),
 \qquad
 |\Phi_r^\zeta(t)|+|R_\zeta(t,\varepsilon)|\le Ce^{-\beta t}.
\]
The constants depend on the fixed dimension, graph, teacher and
initialization; they are not uniform as $r$ grows.  If the
initial-time certificate has $J_r\ne0$, there are a fixed time
$t_\dagger>0$ and constants $a,A>0$ such that, for sufficiently small
positive $\varepsilon$,
\[
 a\varepsilon^r\le
 |L_+^\zeta(t_\dagger,\varepsilon)-L_-^\zeta(t_\dagger,\varepsilon)|
 \le\sup_{t\ge0}|L_+^\zeta-L_-^\zeta|
 \le A\varepsilon^r .
\]
\end{theorem}
\begin{proof}
Put $a_-=\min T_A$, $a_+=\max T_A$,
$t_-=\lambda_{\min}(T_C)$, $t_+=\lambda_{\max}(T_C)$,
$A_*=2\max(c,a_+)$ and $N_*=2\max(c,t_+)$.
Let $b_*=\max_s\|B_s\|_*$ and $b_o=\max_s\|B_s\|_{\rm op}$.
The norm of the complete perturbation is
\[
 k=\max_s\left\{
 \tfrac12\|\ell\|_2\|B_s\|_F,\,
 \|\ell\|_\infty\|B_s\|_{\rm op}\right\}.
\]
Choose $\bar\varepsilon>0$ strictly below the law-positivity radius and
\[
 \min\left\{k^{-1},\
 \frac{a_-}{\|\ell\|_\infty b_*N_*},\
 \frac{t_-}{\|\ell\|_1A_*b_o}\right\}.
\]
A zero denominator is interpreted as $+\infty$.
The effective teachers in \eqref{eq:intrinsicblockflow} are
$T_{A,i}-(\varepsilon/2)\ell_iQ_s$ and
$T_C-(\varepsilon/2)S_sB_s$.  On the box $a_i\le A_*$,
$N_s\preceq N_*I$, their deviations from the original teachers
are bounded by $a_-/2$ and $t_-/2$, respectively.
The scalar logistic equation points inward at
$\min(c,a_-/2)$ and $A_*$.  If $v$ is an eigenvector of $N$ with
eigenvalue $\lambda$, then
\[
 v^\top\dot Nv
 =4\lambda\{v^\top[T_C-(\varepsilon/2)SB_s]v-\lambda\}.
\]
Extremal-eigenvalue Dini derivatives therefore point inward at
$\min(c,t_-/2)I$ and $N_*I$.  A first-exit argument proves invariance
of this compact positive box for all $t\ge0$ and
$|\varepsilon|\le\bar\varepsilon$.  In particular $M_s\succeq mI$
for a fixed $m>0$.

Write $E_s=M_s-T$, $\mu=1-\bar\varepsilon k$,
$\Lambda=1+\bar\varepsilon k$ and
$V_s=\langle E_s,H_sE_s\rangle/2$.  With $G_s=H_sE_s$,
\[
 \dot V_s=-4\tr(M_sG_s^2)\le-\omega V_s,\qquad
 \omega=\frac{8m\mu^2}{\Lambda}.
\]
Consequently
$\|E_s(t)\|_F\le(\Lambda/\mu)^{1/2}\|cI-T\|_F e^{-\omega t/2}$,
uniformly in the coupling.

It remains to justify a uniform Taylor remainder on an unbounded time
interval.  The state variational equation has
\[
 J_s(t)Z=-2\{H_sZ,M_s\}-2\{G_s,Z\}.
\]
For $W_Z=\langle Z,H_sZ\rangle/2$,
\[
 \dot W_Z\le
 \left[-\omega+\frac{8\Lambda}{\mu}\|G_s(t)\|_F\right]W_Z.
\]
The second coefficient is integrable uniformly in $\varepsilon$.
Thus the variational propagator satisfies
$\|U_s(t,u)\|\le C_Ue^{-\omega(t-u)/2}$ for all $t\ge u\ge0$.

The parameter-derivative induction can be written explicitly.
Suppress the arm subscript and define
\[
 \mathcal A_H Y=-2\{HY,T\},\qquad
 \mathcal B_H(Y,Z)=-2\{HY,Z\}.
\]
Then $\dot E=\mathcal A_HE+\mathcal B_H(E,E)$ and
$\partial_\varepsilon H=K$.
Put $X_0=E$ and $X_j=\partial_\varepsilon^jE$ for $j\ge1$.
Leibniz's rule gives $\dot X_j=J_sX_j+R_j$, $X_j(0)=0$, with
\begin{align*}
 R_j={}&\sum_{a=1}^{j-1}\binom ja
             \mathcal B_H(X_a,X_{j-a})+j\mathcal A_KX_{j-1}\\
       &+j\sum_{a=0}^{j-1}\binom{j-1}{a}
             \mathcal B_K(X_a,X_{j-1-a}).
\end{align*}
All factors on the right are $E$ or lower-order derivatives.
The linear and bilinear operator norms are uniformly bounded on the
fixed coupling interval.  For $j=1$, the exponential bound on $E$
gives $\|R_1(t)\|\le C_1e^{-\omega t/2}$.  If the lower-order
derivatives decay as $e^{-\beta t}$, where $\beta=\omega/4$, then
$\|R_j(t)\|\le C_je^{-\beta t}$ for $2\le j\le r+2$.
Using the faster propagator above,
\[
 \|X_j(t)\|\le C_U\int_0^t
     e^{-\omega(t-u)/2}\|R_j(u)\|\,du.
\]
For $j=1$ this is bounded by $Cte^{-\omega t/2}$; for $j\ge2$
it is bounded by $C_je^{-\beta t}/(\omega/2-\beta)$.
Induction therefore proves
\[
 \sup_{|\varepsilon|\le\bar\varepsilon/2,\ t\ge0}
 e^{\beta t}\|\partial_\varepsilon^jE_s(t,\varepsilon)\|_F<\infty.
\]
The constants are uniform in time and coupling for every
$0\le j\le r+2$ at the fixed $r$.
Both losses are quadratic in $E$, and $H$ is affine in the coupling.
A further application of Leibniz's rule gives
\[
 \sup_{|\varepsilon|\le\bar\varepsilon/2,\ t\ge0}
 e^{\beta t}|\partial_\varepsilon^j L_s^\zeta(t,\varepsilon)|
 <\infty,\qquad 0\le j\le r+2,\quad\zeta\in\{G,P\}.
\]

The signed-cycle proof of Theorem~\ref{thm:geo-response} makes all
between-arm derivatives below order $r$, and the derivative of order
$r+1$, vanish at zero coupling for each $t$.
Taylor's formula with the preceding uniform derivative bound proves
the claimed expansion and exponential envelope.
Finally $\Phi_r^\zeta(t)=\overline C_{q,r}^\zeta t^{rq}
+O(t^{rq+1})$ with $\overline C_{q,r}^\zeta\ne0$ gives a nonzero
coefficient at one sufficiently small fixed $t_\dagger$.
Shrinking $\varepsilon_0$ makes the remainder less than half that
signal.  This proves the two-sided order.
\end{proof}

\begin{theorem}[A scalar passive--intervention separation]
\label{thm:geo-channels}
Fix a supplied pair satisfying the nonzero-certificate condition in
Theorem~\ref{thm:geo-uniform}.  The unknown hypothesis is which of
its two input laws is used.  The teacher and alternatives are known.
In both experiments the query complexity is the minimal fixed
deterministic budget $n$, with adaptive query choices within that budget.

In the passive experiment, each query returns
$L_s^\zeta(t_j,\varepsilon)+\sigma_L Z_j$, with independent standard
Gaussian $Z_j$, fixed $\sigma_L>0$, and an adaptively chosen
$t_j\ge0$ on the original trajectory from $W_0=\sqrt cI$.
For error probability at most $\eta$ under each alternative, where
$0<\eta\le1/8$, the optimal number of scalar queries has order
\begin{equation}
 \Theta\!\left(\sigma_L^2\varepsilon^{-2r}\log(1/\eta)\right)
 \qquad(\varepsilon\downarrow0).
 \label{eq:geo-passive}
\end{equation}

In the intervention experiment, choose a fixed $0<\rho<c$.
A query may initialize a fresh factor with
\[
 M(v)=\diag(c\mathbf1+v)\oplus cI,\qquad
 \|v\|_2\le\rho,\qquad W(v)=M(v)^{1/2},
\]
and return the single noisy scalar
\[
 Y=\langle V,G_s(M(v))_C\rangle+\sigma_G Z,
\]
where $V\in\sym(2r-1)$, $\|V\|_F\le1$ and $\sigma_G>0$.
This is a normalized directional measurement of the initial weight gradient:
\[
 G_s(M(v))_C=(\nabla_W L_s^P(W(v)))_{CC}/(2\sqrt c).
\]
The choices of $v,V$ may be adaptive.  In this bounded intervention
class the optimal number of scalar queries instead has order
\begin{equation}
 \Theta\!\left(\sigma_G^2\varepsilon^{-2}\log(1/\eta)\right).
 \label{eq:geo-active}
\end{equation}
Both exponents concern the same two laws and the same teacher; each
observation is scalar.  They do not count input samples, the cost of
training, or the cost of producing a population-gradient measurement.
\end{theorem}
\begin{proof}
For a pair of scalar response functions whose maximum difference is
$D$, the Gaussian chain rule gives, for any adaptive query policy,
\[
 D_{\rm KL}(P_+^{(n)}\|P_-^{(n)})\le\frac{nD^2}{2\sigma^2}.
\]
The policy has the same conditional law under each hypothesis, so
adaptivity adds no extra divergence.  The two-point inequality
$P_+(\mathrm{error})+P_-(\mathrm{error})
\ge\tfrac12 e^{-D_{\rm KL}(P_+\|P_-)}$ \citep{ref45}
implies $n\ge 2\sigma^2D^{-2}\log(1/(4\eta))$.
At a fixed query with signal $d>0$, repeated measurements followed
by a midpoint test have each error at most
$\exp[-nd^2/(8\sigma^2)]$.
For $\eta\le1/8$ the two logarithmic factors are comparable.
The uniform upper signal and one fixed-time lower signal in
Theorem~\ref{thm:geo-uniform} establish \eqref{eq:geo-passive}.

For the intervention, $H_s(cI-T)=cI-T$ and the complete block formula
give
\[
 G_+(M(v))_C-G_-(M(v))_C
   =\frac{\varepsilon}{2}(\ell^\top v)(B_+-B_-).
\]
Thus every allowed scalar query has signal at most
$D_G=\varepsilon\rho\|\ell\|_2\|B_+-B_-\|_F/2$.
Equality is attained by
$v=\rho\ell/\|\ell\|_2$ and
$V=(B_+-B_-)/\|B_+-B_-\|_F$.
The factor is positive because $c+v_i\ge c-\rho>0$.
Applying the same divergence bound and midpoint test proves
\eqref{eq:geo-active}.
\end{proof}

The intervention queries a gradient at a new initialization, while
the passive channel observes the original trajectory.  Both experiments
test the same supplied pair with the same teacher and scalar readouts.
The initialization change removes the cycle penalty in this testing
problem.  Query-dependent statistical difficulty is
well established, including gradient-query models
\citep{joshi2024}.  Here the separation is realized within one
positive-input-law neural construction with exactly matched
diagonal-anchored spectra.

\begin{corollary}[An intervention implemented with finitely many inputs]
\label{cor:geo-finite-input}
Use the smooth laws in Proposition~\ref{prop:geo-smooth}, with their
fixed perturbations $\psi_\pm$, and suppose
$0<\varepsilon R_\psi\le1/2$.
For a supplied pair let
\[
 B_\Delta=B_+-B_-,\quad
 v_*=\rho\ell/\|\ell\|_2,\quad
 V_*=B_\Delta/\|B_\Delta\|_F,\quad
 D_*=\diag(v_*)\oplus0,\qquad 0<\rho<c.
\]
On each independent input form the scalar
\begin{equation}
 Z(x)=\langle V_*,A(x)_C\rangle\langle D_*,A(x)\rangle .
 \label{eq:geo-paired-score}
\end{equation}
This score is also the projection of the difference of the two
per-example weight gradients at $W(v_*)$ and $W(0)$, divided by
$2\sqrt c$, when the same input and label are used in both gradients.
The reduced score $Z(x)$ can be evaluated from the input alone,
without knowing the teacher or a label.  An implementation through
two ordinary per-example gradients uses the same available label
in both evaluations; that label cancels from their difference.
Its means under the two alternatives are
\[
 \mathbb E_\pm Z=\pm d_\varepsilon/2,\qquad
 d_\varepsilon=\frac{\varepsilon\rho\|\ell\|_2
                           \|B_\Delta\|_F}{2},
 \qquad \operatorname{Var}_\pm Z\le\frac32\rho^2.
\]
A median-of-means estimate followed by a zero-threshold test
therefore has error at most $\eta$ under each alternative using
\begin{equation}
 n=O\!\left(
 \frac{\log(1/\eta)}
      {\varepsilon^2\|\ell\|_2^2\|B_\Delta\|_F^2}
 \right),\qquad 0<\eta\le1/8 .
 \label{eq:geo-input-upper}
\end{equation}
For the fixed pair this has the optimal $\varepsilon$ and $\eta$
orders among all tests based on independent raw inputs.
This is a finite-input implementation of pair testing, not a
procedure for recovering an arbitrary unknown graph.
\end{corollary}
\begin{proof}
Write the per-example loss as
$\mathcal L(W;x,y)=\tfrac12(f_W(x)-y)^2$.
Since $\nabla_W\mathcal L=2(f_W-y)A(x)W$ and both initializations
have $W_{CC}=\sqrt cI$ and zero off-diagonal blocks,
\[
 \frac{
 [\nabla_W\mathcal L(W(v_*);x,y)
   -\nabla_W\mathcal L(W(0);x,y)]_{CC}}
 {2\sqrt c}
 =A(x)_C\langle D_*,A(x)\rangle .
\]
Projection onto $V_*$ proves the paired-gradient identity.
The label cancels before averaging, including any shared label
noise.  The moment operator gives
\[
 \mathbb E_s Z
 =\langle V_*,H_s(D_*)_C\rangle
 =\frac{\varepsilon}{2}(\ell^\top v_*)\langle V_*,B_s\rangle .
\]
The two $B_s$ have the same Frobenius norm; consequently
$\langle V_*,B_\pm\rangle=\pm\|B_\Delta\|_F/2$.
Under the product Gaussian, the two factors in
\eqref{eq:geo-paired-score} depend on disjoint coordinate blocks
and have second moments $1$ and $\rho^2$, respectively.
Thus $\mathbb E_{\phi_d}Z^2=\rho^2$.  The density-ratio upper
bound $p_{s,\varepsilon}\le(3/2)\phi_d$ implies the stated
variance bound.

For completeness, divide the inputs into an odd number $b$ of
independent blocks of size $m$.
If $m\ge64(3\rho^2/2)/d_\varepsilon^2$, Chebyshev's inequality
puts each block mean within $d_\varepsilon/4$ of its true mean
with probability at least $3/4$.  The probability that half the
blocks fail is at most $e^{-b/8}$.  Choosing the smallest odd
$b\ge8\log(1/\eta)$ proves \eqref{eq:geo-input-upper}, with
integer rounding absorbed in the asymptotic bound.

For the lower bound put
$D_\psi=\mathbb E_{\phi_d}(\psi_+-\psi_-)^2>0$.
The two perturbations differ because their fourth moments differ.
The one-input divergence satisfies
\[
 D_{\rm KL}(P_{+,\varepsilon}\|P_{-,\varepsilon})
 \le\chi^2(P_{+,\varepsilon}\|P_{-,\varepsilon})
 =\varepsilon^2\mathbb E_{\phi_d}
   \frac{(\psi_+-\psi_-)^2}{1+\varepsilon\psi_-}
 \le\frac{\varepsilon^2D_\psi}{1-\varepsilon R_\psi}.
\]
The two-point inequality used above then requires every fixed
input budget with both errors at most $\eta$ to satisfy
\[
 n\ge\frac{1-\varepsilon R_\psi}{\varepsilon^2D_\psi}
                  \log\frac1{4\eta}.
\]
Labels generated by the same conditional law given the input
under both hypotheses do not change this divergence.  For a fixed
pair and $\eta\le1/8$, the lower and upper orders agree.
\end{proof}

The shared inputs are essential to the exact cancellation in this
implementation.  No independent additive measurement noise is
introduced after the two gradients are computed.
The passive exponent in Theorem~\ref{thm:geo-channels} belongs to
its explicitly defined noisy-loss oracle; it is not a lower bound
for a learner that can inspect raw inputs.  The score above is an
ordinary fourth-moment statistic, which also can be evaluated
directly without computing gradients.

\begin{proposition}[The bounded-teacher continuity boundary]
\label{prop:geo-bounded}
Fix the dimension, a finite time interval, $c>0$, and bounded sets of
positive teachers and perturbations $K_s$.  For
$H_s=\Id+\varepsilon K_s$, the two predictors starting at $cI$ and
their common Gaussian losses differ by $O(\varepsilon)$, uniformly
over those sets for sufficiently small $\varepsilon$.
\end{proposition}
\begin{proof}
The Gaussian reference solutions stay in a common bounded set: their
largest eigenvalues cannot cross the larger of $c$ and the teacher's
largest eigenvalue.  On a slightly larger matrix ball of radius $R$,
the perturbed polynomial vector fields have a common Lipschitz
constant and their differences are bounded by
$4R(R+\|T\|_F)\|H_+-H_-\|_{\rm op}$.
Gronwall's inequality gives an $O(\varepsilon)$ difference from the
reference and closes the ball bootstrap on the fixed time interval.
The common loss is Lipschitz on this ball.  This proves the assertion.
\end{proof}

\section{A logarithmic amplification threshold}
\label{sec:geometry-amplification}

Fix $q,r\ge3$, $c>0$, $a>c$, and a positive diagonal interaction teacher
$T_C$ with distinct entries, none equal to $a$.  Put
\[
 \ell_i=\frac{(-1)^i}{i!(q-1-i)!},\quad i=0,\ldots,q-1,\qquad
 T_{A,i}^{(L)}=a+iL,\qquad
 T_L=\diag(T_A^{(L)})\oplus T_C.
\]
Thus $\ell^\top\mathbf1=\ell^\top T_A^{(L)}=0$ and
$\ell_0=1/(q-1)!$.  For all sufficiently large $L$ the teacher is simple.
Use the fixed moment perturbations \eqref{eq:geo-complete}, with physical
coupling $\varepsilon_L\ge0$.  These are legal positive GH5 laws for all
sufficiently small $\varepsilon_L$; their support and positivity constant
are independent of $L$.  The canonical normalized barycentric vector
would be $\ell/L^{q-1}$, with compensating coupling
$\varepsilon_LL^{q-1}$.  We keep the physical operator normalization
explicit throughout.

\begin{theorem}[Weak moment perturbations with a finite nonlinear limit]
\label{thm:geo-amplification}
Suppose
\[
 \varepsilon_L\longrightarrow0,\qquad
 \varepsilon_L\log L\longrightarrow\kappa\in[0,\infty).
\]
Both learners start at the same full factor $W_0=\sqrt cI$ and train on
the same teacher $T_L$.  Their complete operators converge in norm to
$\Id$ and satisfy Theorem~\ref{thm:geo-equivalence} for each $L$.
Nevertheless, for every compact interval $[t_0,T_0]\subset(0,\infty)$
their interaction predictors converge uniformly to
\begin{align}
 N_s^\kappa(t)
 &=\left[
 c^{-1}e^{-2T_Ct}\exp\!\left(\frac{\kappa\ell_0}{2}B_s\right)e^{-2T_Ct}
 +T_C^{-1}(I-e^{-4T_Ct})
 \right]^{-1}, \label{eq:geo-limit}\\
 N_s^\kappa(0)&=c\exp\!\left(-\frac{\kappa\ell_0}{2}B_s\right).
 \label{eq:geo-seed}
\end{align}
The two own-loss and common-loss differences have the same limit,
\begin{equation}
 \Delta_\kappa(t)
 =\frac12\|N_+^\kappa(t)-T_C\|_F^2
  -\frac12\|N_-^\kappa(t)-T_C\|_F^2. \label{eq:geo-losslimit}
\end{equation}
For $\kappa=0$ this limit vanishes.  If
$D_C=cI-T_C$ and $J_r=\tr[D_C(B_+^r-B_-^r)]\ne0$, then
\begin{equation}
 \Delta_\kappa(0)
  =\frac{c}{r!}\left(-\frac{\kappa\ell_0}{2}\right)^rJ_r
     +O(\kappa^{r+2}). \label{eq:geo-limitcontrast}
\end{equation}
Thus every sufficiently small fixed $\kappa>0$ gives nonzero limiting
separation on some positive time interval.
More quantitatively, set
\[
 \tau_L=\frac{(\log L)^2}{L},\qquad
 b_L=L^2e^{-2(\log L)^2},\qquad
 \eta_L=|\varepsilon_L\log L-\kappa|+\varepsilon_L+\tau_L+b_L.
\]
For every fixed $T_0>0$ there is a constant $C$, independent of $L$ and
of the arm, such that for all sufficiently large $L$,
\begin{align}
 \sup_{\tau_L\le t\le T_0}
       \|N_s(t)-N_s^\kappa(t)\|_{\rm op}&\le C\eta_L,
       \label{eq:geo-quant-predictor}\\
 \sup_{\tau_L\le t\le T_0}
       |(L_+^\zeta-L_-^\zeta)(t)-\Delta_\kappa(t)|
       &\le C\eta_L,\quad \zeta\in\{G,P\}.
       \label{eq:geo-quant-loss}
\end{align}
In particular, $\varepsilon_L=\kappa/\log L$ gives an
$O(1/\log L+(\log L)^2/L)$ error bound.  The constants may depend on
the fixed teacher, graph, initialization, $\kappa$, and $T_0$.
\end{theorem}

\begin{lemma}[Uniform initial-layer estimates]
\label{lem:geo-layer}
Under the assumptions of Theorem~\ref{thm:geo-amplification}, both arms
have, uniformly on $[0,\tau_L]$, bounded $N_s$ and bounded exposure
$u_s=\varepsilon_L\int_0^t S_s(v)\,dv$.  At the end of this interval,
\begin{align*}
 |a_0(\tau_L)-c|&\le C\tau_L,\\
 |a_i(\tau_L)-(a+iL)|&\le C(\varepsilon_L+b_L),\quad i\ge1,\\
 |u_s(\tau_L)-\kappa\ell_0/4|
   &\le C(|\varepsilon_L\log L-\kappa|+\varepsilon_L),\\
 \|N_s(\tau_L)-c e^{-\kappa\ell_0 B_s/2}\|_{\rm op}
   &\le C(|\varepsilon_L\log L-\kappa|+\varepsilon_L+\tau_L).
\end{align*}
\end{lemma}
\begin{proof}
Write $a_i(t)$ for auxiliary states, $S=\sum_i\ell_i a_i$,
$Q=\tr(B_sN)$ and $u(t)=\varepsilon_L\int_0^tS(v)\,dv$.
The exact block equations are
\[
 \dot a_i=-4a_i(a_i-T_{A,i}^{(L)})
          -2\varepsilon_L\ell_i a_iQ,\qquad
 \dot N=-4N^2+2\{T_C,N\}-\varepsilon_LS\{B_s,N\}.
\]
Positivity permits integration of the logarithmic auxiliary equations.
Using $\ell^\top T_A^{(L)}=0$ gives the exact identity
\begin{equation}
 u(t)=-\frac{\varepsilon_L}{4}
          \sum_i\ell_i\log\frac{a_i(t)}c
       -\frac{\varepsilon_L^2}{2}\|\ell\|_2^2
          \int_0^tQ(v)\,dv. \label{eq:geo-exposure}
\end{equation}

Stop the solution when $\|N\|_{\rm op}=K$, for a fixed $K$ chosen below.
On this interval, $|Q|\le\|B_s\|_*K$.  Set
$h=(\varepsilon_L/2)\|\ell\|_\infty\max_s\|B_s\|_*K$.
Scalar comparison places each $a_i$ between
\[
 b_R(t)=\frac{R}{1+(R/c-1)e^{-4Rt}},
 \qquad R=a+iL-h\quad\hbox{and}\quad R=a+iL+h.
\]
For fixed $K$ and sufficiently large $L$, all these targets exceed $c$.
They also satisfy $R\le C'L$, with $C'$ independent of $K$ once $L$ is
large enough.  Thus
\[
 0\le\log(a_i(t)/c)\le\log(C'L/c),\qquad
 |u(t)|\le \frac{\varepsilon_L}{4}\|\ell\|_1\log(C'L/c)
                   +C''\varepsilon_L^2K\tau_L.
\]
Choose $\bar u>1+\|\ell\|_1(\kappa+1)/4$ and then
$K>2c\exp(2\bar u\max_s\|B_s\|_{\rm op})$.  For this fixed $K$,
the preceding inequality gives $|u|\le\bar u$ when $L$ is sufficiently
large.  This order of choices avoids a circular bootstrap constant.

The change of variables $Z=e^{uB_s}Ne^{uB_s}$ cancels the entire
$\varepsilon_LS$ term.  On the stopped interval its derivative is
$e^{uB_s}[-4N^2+2\{T_C,N\}]e^{uB_s}$, with norm at most
\[
 e^{2\bar u\|B_s\|_{\rm op}}
       (4K^2+4\|T_C\|_{\rm op}K).
\]
Hence $\|Z(t)-cI\|_{\rm op}\le C_K\tau_L$.
For large $L$, the bound
$\|N(t)\|_{\rm op}\le e^{2\bar u\|B_s\|_{\rm op}}(c+C_K\tau_L)<K$
prevents the stopping event.

For $i\ge1$, the explicit logistic formula gives, for all
$t\ge\tau_L$ on any interval where the same bound on $Q$ holds,
\[
 |a_i(t)-(a+iL)|
 \le h+C L^2e^{-2Lt}\le C(\varepsilon_L+b_L).
\]
For $i=0$, boundedness of the two logistic targets gives
$|a_0(t)-c|\le Ct$ in the initial layer.
Since $\sum_{i\ge1}\ell_i=-\ell_0$,
\[
 \sum_i\ell_i\log(a_i(\tau_L)/c)
 =-\ell_0\log L+\sum_{i\ge1}\ell_i\log(i/c)
        +O\!\left(\tau_L+L^{-1}(1+\varepsilon_L+b_L)\right).
\]
The integral term in \eqref{eq:geo-exposure} has magnitude at most
$C\varepsilon_L^2\tau_L$.  This proves the exposure bound.
The matrix exponential is Lipschitz on the fixed bounded interval
of exponents.  Combining it with $\|Z(\tau_L)-cI\|\le C\tau_L$
proves the final predictor estimate.
\end{proof}

\begin{lemma}[Uniform continuation and loss estimates]
\label{lem:geo-continuation}
For every fixed $T_0>0$, the constants in the preceding lemma can be
enlarged so that $N_s$ remains uniformly bounded on $[0,T_0]$ and
\[
 \sup_{\tau_L\le t\le T_0}|\varepsilon_LS_s(t)|
       \le C\varepsilon_L,\qquad
 \sup_{\tau_L\le t\le T_0}
 |a_{s,i}(t)-(a+iL)|\le C(\varepsilon_L+b_L)\quad(i\ge1).
\]
Moreover \eqref{eq:geo-quant-predictor} and
\eqref{eq:geo-quant-loss} hold.
\end{lemma}
\begin{proof}
The Gaussian equation $\dot N=-4N^2+2\{T_C,N\}$ with initial state
$c e^{-\kappa\ell_0B_s/2}$ has its largest eigenvalue bounded by
the larger of the initial largest eigenvalue and $\|T_C\|_{\rm op}$.
Call a common bound $K_0$.  Stop the actual solution after $\tau_L$
if $\|N_s\|_{\rm op}$ reaches a fixed $K_1>K_0+1$, enlarging $K_1$
to cover the entire uniform initial-layer bound if needed.  On this stopped interval,
the preceding scalar comparisons apply with $K_1$ in place of $K$.
Their explicit formula proves the displayed fast-coordinate estimate.
The slow coordinate remains bounded above and below by logistic
solutions with targets $a\pm C\varepsilon_L$.
Since $\ell^\top T_A^{(L)}=0$,
\[
 S_s=\ell_0(a_{s,0}-a)+
                \sum_{i\ge1}\ell_i(a_{s,i}-(a+iL))
       =O(1)+O(\varepsilon_L+b_L).
\]
Thus $|\varepsilon_LS_s|\le C\varepsilon_L$ for large $L$.

On the fixed matrix ball the Gaussian vector field has a Lipschitz
constant $C_1$.  Its perturbing term has norm at most
$2\varepsilon_L|S_s|\|B_s\|_{\rm op}K_1\le C_2\varepsilon_L$.
The initial mismatch at $\tau_L$, including the $O(\tau_L)$
advance of the limiting trajectory, is at most $C_3\eta_L$.
Gronwall therefore gives
\[
 \sup_{\tau_L\le t\le T_0}
 \|N_s(t)-N_s^\kappa(t)\|_{\rm op}
 \le e^{C_1T_0}(C_3\eta_L+C_2T_0\varepsilon_L).
\]
This tends to zero and prevents the stopping event.  Together with
the initial-layer bound it also bounds $Q_s$ on $[0,T_0]$.

Let $\bar a$ solve $\dot{\bar a}=-4\bar a(\bar a-a)$, $\bar a(0)=c$.
The slow auxiliary forcing is bounded by $C\varepsilon_L$ on
$[0,T_0]$, so scalar Gronwall gives
$\sup|a_{s,0}-\bar a|\le C\varepsilon_L$.
The fast-coordinate squared-error sum is bounded by
$C(\varepsilon_L+b_L)^2$ on $[\tau_L,T_0]$.
The interaction squared loss is Lipschitz on the fixed ball.
Consequently
\[
 \sup_{\tau_L\le t\le T_0}
 \left|L_s^G(t)-
 \left\{\tfrac12(\bar a(t)-a)^2+
             \tfrac12\|N_s^\kappa(t)-T_C\|_F^2\right\}\right|
 \le C\eta_L.
\]
Finally $|L_s^P-L_s^G|=|\varepsilon_LS_sQ_s|/2
\le C\varepsilon_L$.  Subtraction between arms cancels the common
slow-coordinate limit and proves \eqref{eq:geo-quant-loss}.
\end{proof}

\begin{proof}[Proof of Theorem~\ref{thm:geo-amplification}]
The two lemmas prove the effective seed, both uniform convergences,
and the quantitative bounds.  The inverse of the limiting predictor
satisfies $\dot R=4I-2\{T_C,R\}$, which integrates to
\eqref{eq:geo-limit}.  Complete spectral equivalence holds at each
$L$ by Theorem~\ref{thm:geo-equivalence}; the fixed operators $K_s$
are bounded, so $H_s\to\Id$.
At the effective initial state, conjugacy of $B_+$ and $B_-$ makes
$\tr(N_+^2)=\tr(N_-^2)$ and $\tr(N_+)=\tr(N_-)$, leaving
\[
 \Delta_\kappa(0)=
 c\,\tr\!\left[D_C
 \left(e^{-\kappa\ell_0B_+/2}-e^{-\kappa\ell_0B_-/2}\right)\right].
\]
Shorter signed closed walks agree.  The first differing term is the
length-$r$ term, and parity excludes length $r+1$, proving
\eqref{eq:geo-limitcontrast}.  Continuity in positive time finishes the
proof.
\end{proof}

\paragraph{Scope of the threshold.}
The threshold $\varepsilon_L\log L$ is exact for this specified teacher
family, not a universal condition-number law for neural networks.
Teacher norm and initial loss diverge with $L$, so the result does not
contradict continuous dependence for bounded teachers.  Initialization
is fixed and full rank; the singularity is in teacher scale and moment
coupling.  The effective initial state in \eqref{eq:geo-seed} is produced
by the original dynamics in a vanishing initial time interval, not
imposed as a reset or substituted training objective.  For every finite
$L$, the true initial state remains $cI$.  When $\kappa>0$ and $B_s\ne0$,
the limiting state has a discontinuity at zero relative to this
true initial condition.  Uniform convergence is asserted only after
the shrinking layer $[0,\tau_L]$.

The direct predictor flow $\dot M=-4H_s(M-T_L)$ is a useful control.
Here $H_s(cI-T_L)=cI-T_L$, so it equals
$T_L+e^{-4t}(cI-T_L)$ under both laws for every $L$.  Its between-arm loss
gap is identically zero.

\section{What a scalar learning curve identifies}
\label{sec:inverse}

The preceding results compare spectral information available before
training.  We now ask a different question: what information is present
in the scalar history after nonlinear learning has occurred?
Consider the Gaussian factorized flow
\[
 \dot N=-4N^2+2\{T,N\},\qquad N(0)=M_0\succ0,
 \qquad T=\diag(t_1,\ldots,t_d)\succ0.
\]
We first take the teacher and time scale as known;
Section~\ref{sec:inverse-blind} removes the teacher assumption.
Put $C=M_0^{-1}-T^{-1}$ and $D_s=e^{-2Ts}$.
The established inverse-matrix solution is
\begin{equation}
 N_C(s)=[T^{-1}+D_sCD_s]^{-1},\qquad
 L_C(s)=\tfrac12\|N_C(s)-T\|_F^2.
 \label{eq:inverse-riccati}
\end{equation}
The question is not how to solve this Riccati equation, but whether its
scalar loss identifies the initial matrix.

\begin{theorem}[The exact scalar tail is a complete invariant]
\label{thm:inverse-global}
Suppose $t_1,\ldots,t_d$ are linearly independent over $\mathbb Q$.
For any $C,\widetilde C$ such that
$T^{-1}+C\succ0$ and $T^{-1}+\widetilde C\succ0$, the following are equivalent:
\begin{enumerate}
\item $L_C(s)=L_{\widetilde C}(s)$ for all sufficiently large $s$;
\item $\widetilde C=SCS$ for a diagonal matrix
$S=\diag(\epsilon_1,\ldots,\epsilon_d)$, $\epsilon_i\in\{-1,1\}$;
\item $\widetilde M_0=SM_0S$ for such a matrix $S$.
\end{enumerate}
There is no density or connectedness assumption on the off-diagonal
support of $C$.
\end{theorem}
\begin{proof}
The inverse state satisfies $\dot R=4I-2\{T,R\}$, giving
\eqref{eq:inverse-riccati}.  Define
$Z=T^{1/2}D_sCD_sT^{1/2}$.  At sufficiently late times the entrywise
absolute matrix has $\||Z|\|_\infty<1$.  The Neumann expansion therefore
converges absolutely even after expanding each trace into paths:
\begin{equation}
 L_C(s)=\frac12\sum_{m,n\ge1}(-1)^{m+n}
                  \tr(TZ^mTZ^n).
 \label{eq:inverse-series}
\end{equation}
Every closed path has a vertex occupation vector
$\nu\in\mathbb N^d$, $|\nu|\ge2$, and exponential rate
$4\nu\cdot t$.  Absolute convergence permits regrouping to obtain
\[
 L_C(s)=\sum_{|\nu|\ge2}a_\nu(C)e^{-4(\nu\cdot t)s}.
\]
The positive rates form a locally finite set; rational independence
makes different occupation vectors have different rates.  These
coefficients are unique.  Indeed, in a difference of two such series,
take the least rate with a nonzero coefficient, multiply by its inverse
exponential and send $s$ to infinity.  Absolute convergence at one fixed
late time supplies a summable dominating sequence, so every larger-rate
term vanishes.  The nonzero leading coefficient is a contradiction.

Expanding the degree-two and degree-three terms gives
\begin{align}
 a_{2e_i}&=\tfrac12t_i^4C_{ii}^2,&
 a_{3e_i}&=-t_i^5C_{ii}^3,&
 a_{e_i+e_j}&=t_i^2t_j^2C_{ij}^2\quad(i<j).
 \label{eq:inverse-low}
\end{align}
Thus every diagonal entry, every off-diagonal magnitude and the support
graph $G_C$ are known.

Let $V\subset[d]$, $|V|=k\ge3$.  If $\operatorname{Ham}(G_C[V])$ denotes
the undirected Hamiltonian cycles, counted as unoriented edge sets,
then the coefficient of occupation vector $\mathbf1_V$ is
\begin{equation}
 a_{\mathbf1_V}
 =(-1)^k
 \underbrace{\left(\prod_{v\in V}t_v\right)
             \left(\sum_{\substack{a,b\in V\\a\ne b}}t_at_b\right)}_{K_V>0}
 \sum_{H\in\operatorname{Ham}(G_C[V])}
              \prod_{\{i,j\}\in E(H)}C_{ij}.
 \label{eq:inverse-cycle}
\end{equation}
To check the factor, each cyclic path uses every vertex once.  Its
edge factors supply $\prod_{v\in V}t_v$; the two marked positions of
the external $T$ factors supply $t_at_b$.  Summing the initial vertex
and the split $m=1,\ldots,k-1$ supplies every ordered distinct pair.
The two path orientations cancel the prefactor $1/2$ in
\eqref{eq:inverse-series}.  No diagonal stay or repeated vertex has
this occupation vector, and no other vector shares its rate.

If $G_C[V]$ is a chordless cycle, its cycle is the only Hamiltonian
edge set.  Equation~\eqref{eq:inverse-cycle} determines its signed edge
product.  A cycle containing a chord decomposes into two shorter cycles,
so the chordless products determine every cycle-sign invariant.
This last reconstruction step is classical signed-graph and
principal-minor machinery \citep{ref59,urschel2017,brunel2018}.
For completeness, the ratio of the two edge signs then has product $+1$
around every cycle.  Choose a root in each connected component and
propagate vertex signs along paths from that root.  The cycle condition
makes these signs path-independent and gives
$\widetilde C_{ij}=\epsilon_i\epsilon_j C_{ij}$.  Diagonal entries already
agree, proving the second condition.  It is equivalent to the third by
inversion.  Conversely $S$ commutes with $T$ and $D_s$, so
$N_{SCS}(s)=SN_C(s)S$, which leaves the loss unchanged.
\end{proof}

The theorem uses an exact continuum of observations with unlimited
precision.  Analyticity implies that exact equality on any nonempty
open time interval also suffices.  This implication is not a stable
analytic-continuation algorithm.  Rational independence is a sufficient
condition for the coefficient proof, not a claimed necessary condition
for identifiability.

\begin{theorem}[A minimal number of exact scalar observations for generic local recovery]
\label{thm:inverse-finite}
Let $p=d(d+1)/2$ and suppose all unordered sums $t_i+t_j$, $i\le j$,
are distinct.  Fix any $p$ distinct nonnegative times
$s_1,\ldots,s_p$.  On the domain
$\mathcal D=\{C:T^{-1}+C\succ0\}$, the map
\[
 F(C)=(L_C(s_1),\ldots,L_C(s_p))
\]
has an invertible derivative on an open dense subset.  At each such
point it has a local analytic inverse.  Fewer than $p$ real scalar
observations cannot have a continuous local inverse on an open subset
of the full $p$-dimensional parameter space.
\end{theorem}
\begin{proof}
The matrix inside the inverse in \eqref{eq:inverse-riccati} is positive:
it equals
$D_s(T^{-1}+C)D_s+T^{-1}(I-D_s^2)$.
Thus $F$ is real analytic throughout the open convex set $\mathcal D$.
Near $C=0$,
\[
 L_C(s)=\sum_{i\le j}w_{ij}C_{ij}^2e^{-\lambda_{ij}s}
                +O(\|C\|^3),\quad
 \lambda_{ij}=4(t_i+t_j),\quad
 w_{ii}=\tfrac12t_i^4,\quad w_{ij}=t_i^2t_j^2\ (i<j).
\]
Choose a symmetric $\overline C$ whose independent entries are nonzero.
For sufficiently small $\delta\ne0$, $\delta\overline C\in\mathcal D$ and
\begin{align}
 \det DF(\delta\overline C)
 &=(2\delta)^p
       \left(\prod_{i\le j}w_{ij}\overline C_{ij}\right)
       \det[e^{-\lambda_{ij}s_k}]_{k,(i,j)}
       +O(\delta^{p+1}). \label{eq:inverse-determinant}
\end{align}
The exponential evaluation determinant is nonzero.  One elementary
proof is induction on the number of distinct rates: divide a linear
combination by its slowest exponential and differentiate.  Rolle's
theorem and the induction hypothesis show that a nonzero combination
of $p$ exponentials has at most $p-1$ distinct zeros.
Therefore $\det DF$ is not identically zero.  A nonzero real-analytic
function on a connected domain has a zero set with empty interior,
and continuity makes its nonzero set open.  The inverse function
theorem proves local analytic recovery.  The lower bound follows from
invariance of dimension: an open subset of $\mathbb R^p$ cannot be
continuously embedded into $\mathbb R^m$ for $m<p$.
\end{proof}

For an invertible observation Jacobian, the local noise amplification
depends on $\sigma_{\min}(DF(C))^{-1}$.  Its exact derivative can be
computed without differentiating a numerical ODE:
\begin{equation}
 dL_C(s)[E]=-\tr[D_sN_C(s)(N_C(s)-T)N_C(s)D_sE].
 \label{eq:inverse-jacobian}
\end{equation}
Generic local identifiability gives no uniform lower bound on this
singular value and no guarantee for a globally initialized solver.
In particular $DF(0)=0$, and rates or observation times can approach
collisions.  The local theorem and nonlinear output observability
\citep{hermann1977} address related questions; the present statement
computes the observation threshold for this particular loss map.

\begin{corollary}[Reading the interaction left by the initial layer]
\label{cor:inverse-seed}
In Theorem~\ref{thm:geo-amplification}, suppose the interaction-teacher
entries are rationally independent, and the known constants satisfy
$\kappa\ell_0\ne0$.  After subtracting the common known slow auxiliary
loss from the limiting curve, that curve identifies $B_s$ up to
diagonal sign conjugation.  Indeed
\[
 B_s=-\frac{2}{\kappa\ell_0}\log(N_s^\kappa(0)/c).
\]
The finite-observation local theorem also transfers to arbitrary symmetric
interaction matrices through this analytic change of variables.
\end{corollary}
\begin{proof}
The symmetric exponential and the principal logarithm are inverse
analytic maps between symmetric and positive-definite matrices.
They commute with orthogonal conjugation.  Apply
Theorems~\ref{thm:inverse-global} and \ref{thm:inverse-finite}.
\end{proof}

The known slow auxiliary loss in this corollary is necessary: the
observation must be the interaction loss, or a full limiting loss from
which that common term is known.  It is also a statement about the
derived limit, not an exact recovery theorem for a finite-$L$ coupled
trajectory.  The numerical teacher used for the amplification example
has resonant entries and is not presented as a test of the global
inverse theorem.

\subsection{Joint identification of the teacher and initial state}
\label{sec:inverse-blind}

The known-teacher assumption can be removed on an explicitly described
set of initial states.  The time unit and the absolute scale of the loss
are assumed known.  In a teacher eigenbasis put
$C=M_0^{-1}-T^{-1}$ as before.

\begin{theorem}[An unknown teacher is identifiable from the scalar curve]
\label{thm:inverse-blind}
Consider the class of pairs $(T,M_0)$ with $T,M_0\succ0$, rationally
independent eigenvalues of $T$, and $C_{ii}\ne0$ for every $i$ in
a teacher eigenbasis.  Two members of this class have the same exact
loss curve on a nonempty open time interval if and only if their
dimensions agree and they are simultaneously orthogonally conjugate.
Thus the curve identifies the dimension, the teacher eigenvalues and
the initial state modulo this unavoidable symmetry.
For each fixed admissible teacher, the condition on $C$ is open dense
and has full Lebesgue measure in $\mathcal D$.
It imposes no condition on the off-diagonal support.
\end{theorem}
\begin{proof}
Analyticity extends equality on the interval to all positive times.
For either candidate the absolutely convergent late-time expansion
has rates
\[
  4\nu\cdot t,\qquad \nu\in\mathbb N^d,\quad |\nu|\ge2.
\]
The union of the two candidates' rate sets is locally finite.
The least-rate argument in Theorem~\ref{thm:inverse-global} therefore
shows that the nonzero exponent support $\mathcal S$ of the scalar
curve is unique even when the candidate teachers differ.

Write $b_i=4t_i$ and let $\mathcal Q$ be the additive monoid generated
by $\mathcal S$, including zero.  The two coefficients
in~\eqref{eq:inverse-low} imply
$2b_i,3b_i\in\mathcal S$, because $C_{ii}\ne0$ and rational
independence prevents collisions with other occupation vectors.
Consequently
\[
 \mathcal G:=\mathcal Q-\mathcal Q=\bigoplus_{i=1}^d\mathbb Zb_i.
\]
The following set is determined entirely by the observed support:
\[
 \operatorname{Sat}_{\mathcal G}(\mathcal Q)
 =\{g\in\mathcal G:ng\in\mathcal Q
                         \text{ for some integer }n\ge1\}
 =\bigoplus_{i=1}^d\mathbb N b_i .
\]
For the forward inclusion, express $g$ in its unique integer
coordinates in the rationally independent basis $b_i$.
Since every element of $\mathcal Q$ has nonnegative coordinates,
so must $g$ when $ng\in\mathcal Q$.
Conversely, for any $\nu_i\ge0$,
$2\sum_i\nu_i b_i=\sum_i\nu_i(2b_i)\in\mathcal Q$.
The nonzero elements of this saturated monoid that cannot be
written as a sum of two nonzero elements are exactly $b_1,\ldots,b_d$.
The curve therefore determines $d$ and the unordered teacher spectrum.

Order that spectrum, choose teacher eigenbases, and apply
Theorem~\ref{thm:inverse-global}.  The remaining freedom is diagonal
sign conjugation in these bases, which is simultaneous orthogonal
conjugation of the original pair.
Such conjugation preserves the dynamics and loss, proving the converse.
The excluded conditions $C_{ii}=0$ are a finite collection of
proper hyperplanes intersected with the open set $\mathcal D$.
\end{proof}

The monoid here is generated by the observed support; the support
itself need not be closed under addition.
No assumption that every exponential coefficient is nonzero is needed.
There are real obstructions outside the stated class:
$C=0$ gives the zero curve for every teacher, and appending an
uncoupled coordinate with $C_{ii}=0$ leaves the loss unchanged.
The theorem fixes the time, loss and gradient-flow normalizations;
joint calibration from differently normalized measurements is a
separate question.
Recovering all exact exponents is an identifiability argument,
not a stable spectral estimation procedure.
Finite exponential-sum estimation and its noise sensitivity are
established topics in Prony methods \citep{potts2010}.

\begin{theorem}[The minimal local observation count with an unknown teacher]
\label{thm:inverse-blind-finite}
Let $p=d(d+1)/2$ and $m=p+d$.
Suppose all unordered sums $t_i+t_j$, $i\le j$, are distinct.
For any $m$ distinct nonnegative times, the observation map
\[
 (t,C)\longmapsto
       (L_{t,C}(s_1),\ldots,L_{t,C}(s_m))
\]
has an invertible joint derivative at an open dense, full-measure
set of $C\in\mathcal D$ for each such fixed $t$.
It has a local analytic inverse there, with both $t$ and $C$ allowed
to vary.  Fewer than $m$ real scalar observations cannot have a
continuous local inverse on an open $m$-dimensional parameter set.
\end{theorem}
\begin{proof}
For fixed small $\delta>0$, use positive coordinates
\[
 C_{ij}=\delta\sqrt{\alpha_{ij}/w_{ij}(t)},\qquad
 \alpha_{ij}>0,
\]
where $w_{ii}=t_i^4/2$ and $w_{ij}=t_i^2t_j^2$ for $i<j$.
This is a local analytic change of variables from $(\alpha,t)$
to $(C,t)$.  For fixed positive $\alpha$ the loss expansion and
its first derivatives give
\[
 \delta^{-2}L_{t,C}(s)
 =\sum_{i\le j}\alpha_{ij}e^{-\lambda_{ij}s}+O(\delta),
 \qquad \lambda_{ij}=4(t_i+t_j).
\]
The leading expression is linear in $\alpha$ and extends to its
boundary.  At the limiting choice $\alpha_{ii}=1$ and
$\alpha_{ij}=0$ for $i<j$, its Jacobian has the $p$ columns
$e^{-\lambda_{ij}s}$ and the $d$ columns
$-8s e^{-8t_i s}$.
These functions form an extended exponential Chebyshev system:
the real rate $8t_i$ has multiplicity two, and the other rates
have multiplicity one.  Its evaluation matrix at any $m$ distinct
times is invertible.  The zero count follows by successively
dividing by an exponential and differentiating; repeated rates
produce polynomial factors and Rolle's theorem gives at most
$m-1$ distinct zeros for a nonzero combination.

This boundary calculation concerns only the leading map, not the
square-root coordinate change.  By continuity, first fix strictly
positive off-diagonal $\alpha_{ij}$ sufficiently close to zero
that the leading determinant remains nonzero.
For this fixed interior choice, choose $\delta>0$ sufficiently small
that the true joint derivative is invertible and $T^{-1}+C\succ0$.
The analytic coordinate change is applied only at that interior point.
For each fixed $t$ this gives one point where the determinant
is nonzero.  That determinant is real analytic in $C$ on the
connected domain $\mathcal D$.
Its nonzero set is open dense and its zero set has Lebesgue measure
zero.  The analytic inverse function theorem proves joint local
recovery.  Invariance of dimension proves the lower bound.
The finite sign and permutation ambiguities do not alter the
local dimension at these regular points.
\end{proof}

This local count concerns the quotient by simultaneous rotations:
after choosing a teacher eigenbasis the continuous parameters are
its $d$ eigenvalues and the $p$ entries of $C$.
It does not recover an externally fixed orientation from a scalar.
The smallest joint singular value, rather than the parameter count,
controls local noise amplification.

\begin{proposition}[A finite global observation bound]
\label{prop:inverse-global-finite}
For a fixed rationally independent teacher in dimension $d$,
any $5^d$ distinct nonnegative exact loss observations identify
every admissible $C$ up to diagonal sign conjugation.
If the teacher and dimension are unknown but the dimension is at
most $D$, any $9^D$ such observations identify every member of
the class in Theorem~\ref{thm:inverse-blind} up to simultaneous
orthogonal conjugation.
These sufficient bounds are not asserted to be sharp.
\end{proposition}
\begin{proof}
Set $a_i=t_i^{-1}$ and $x_i=e^{-4t_i s}$.  The determinant
\[
 P(a,x)=\det(\diag(a)+\diag(\sqrt x)C\diag(\sqrt x))
       =\sum_{S\subseteq[d]}\det C_{S,S}
                       \prod_{i\in S}x_i\prod_{i\notin S}a_i
\]
is multiaffine in $x$, with $\det C_{\varnothing,\varnothing}=1$.
Let
\[
 U=\left.\partial_zP(a+z\mathbf1,x)\right|_{z=0},\qquad
 V=\left.\partial_z^2P(a+z\mathbf1,x)\right|_{z=0},\qquad
 W=\sum_i t_i\,\partial_{a_i}P(a,x),
\]
where $t$ is held fixed in the last differentiation and
$a_i=t_i^{-1}$ is substituted afterwards.
The determinant derivative identities give
$\tr(N^2)=(U^2-PV)/P^2$ and $\tr(NT)=W/P$.
Hence
\[
 L_{t,C}(s)=\frac{Q(x)}{P(a,x)^2},\qquad
 Q=\tfrac12\{U^2-PV-2PW+\tr(T^2)P^2\},
\]
where $Q$ has degree at most two in each coordinate.
The denominator is positive throughout $s\ge0$.

For two states with the same teacher, clearing denominators
in their loss difference gives a polynomial of degree at most
four in each $x_i$.  On the trajectory it is therefore an
exponential polynomial with at most $5^d$ distinct real rates.
For two teachers of dimensions $d,\widetilde d\le D$,
use separate variables $x_i=e^{-4t_i s}$ and
$\widetilde x_j=e^{-4\widetilde t_j s}$.
The numerator has degree at most two in each variable, so it
has at most $3^{d+\widetilde d}\le9^D$ rates.
Coincident rates only reduce this count.
The exponential zero bound used above implies that the asserted
number of equal observations forces equality of the entire curves.
Apply Theorem~\ref{thm:inverse-global} or
Theorem~\ref{thm:inverse-blind}, respectively.
\end{proof}

General finite-experiment identifiability is established:
\citet{sontag2002} proves that $2k+1$ generic experiments preserve
the distinguishability of an analytic $k$-parameter response.
Applied to this analytic loss map in fixed dimension, that result
already gives smaller generic global observation counts,
$2p+1$ for a known teacher and $2(p+d)+1$ for an unknown teacher.
The elementary bounds above concern \emph{every} set of distinct
times.  The model-specific conclusions are the exact equivalence
class, the blind spectrum recovery, and the dimension-minimal
local count for every distinct-time design.
None of these results supplies a noise-uniform inversion algorithm.

\subsection{Exact identifiability can have an arbitrarily weak statistical signal}
\label{sec:inverse-statistical}

We make the distinction between uniqueness and measurement explicit.
This experiment observes a known pair of Gaussian learning curves with
additive measurement noise.  Its inputs are scalar losses, not samples
from the input laws of Sections~\ref{sec:geometry}--\ref{sec:geometry-amplification}.

\begin{theorem}[Sharp cycle-length exponent for noisy loss queries]
\label{thm:inverse-noise}
Let $d=r\ge3$ and $T=\diag(t_1,\ldots,t_r)\succ0$.  Let $A_\pm$ have
zero diagonal, the common support of one chordless $r$-cycle, positive
edge magnitudes $a_e$, and opposite cycle sign products.  Put
$C_\pm=\delta A_\pm$ and $P=\prod_e a_e$.  Define
\[
 R=\|T^{1/2}|A_\pm|T^{1/2}\|_\infty,\quad
 \rho=\delta_0R<1.
\]
For all $0<\delta\le\delta_0$, both initial states are admissible and
\begin{equation}
 \sup_{s\ge0}|L_+(s)-L_-(s)|\le K_{\rm up}\delta^r,
 \qquad
 K_{\rm up}=r\,t_{\max}^2R^r
 \left[\frac{r-1}{1-\rho^2}
       +\frac{2\rho^2}{(1-\rho^2)^2}\right].
 \label{eq:inverse-noise-upper}
\end{equation}
For all sufficiently small $\delta$ the gap at zero also satisfies
$|L_+(0)-L_-(0)|\ge c_0\delta^r$ for a fixed $c_0>0$.

Suppose the supplied alternatives are observed through
$Y_j=L_\theta(s_j)+\sigma Z_j$, where $\theta\in\{+,-\}$,
$\sigma>0$, the $Z_j$ are independent standard Gaussians, and the
nonnegative observation times may be chosen adaptively.
If both testing errors are at most $\eta<1/4$, necessarily
\begin{equation}
 n\ge
 \frac{2\sigma^2}{K_{\rm up}^2\delta^{2r}}
             \log\frac1{4\eta}.
 \label{eq:inverse-noise-lower}
\end{equation}
Conversely, repeated measurements at zero and a midpoint test attain
error at most $\eta$ with
$n=O(\sigma^2\delta^{-2r}\log(1/\eta))$ as $\delta\to0$.
Thus adaptive selection of observation times cannot improve the
$\delta^{-2r}$ exponent for this supplied pair.
\end{theorem}
\begin{proof}
The norm condition gives
$I+\delta T^{1/2}A_\pm T^{1/2}\succ0$ by its spectral-radius bound,
and therefore $T^{-1}+\delta A_\pm\succ0$.
For every $s\ge0$, the absolute Neumann matrix obeys
$\||Z_\pm(s)|\|_\infty\le\delta R<1$.
In a closed-walk term of \eqref{eq:inverse-series}, changing one edge
sign can matter only if that edge is used oddly.  The odd edge set of
a closed walk is an even subgraph.  On a single chordless cycle it is
empty or the whole cycle, so every differing term has degree $r+2m$.
The zero diagonal excludes additional one-step stays.
At degree $k$, the sum has $k-1$ choices of the two positive path
lengths.  Bounding both traces in the difference gives
\[
 |\Delta L(s)|\le r t_{\max}^2
        \sum_{m\ge0}(r+2m-1)(\delta R)^{r+2m}.
\]
The geometric sum is bounded by \eqref{eq:inverse-noise-upper}.
At zero, the explicit cycle coefficient gives
\[
 \Delta L(0)=2(-1)^r
       \left(\prod_{i=1}^r t_i\right)
       \left(\sum_{a\ne b}t_at_b\right)P\,\delta^r
       +O(\delta^{r+2}),
\]
whose leading coefficient is nonzero.  This proves the lower gap bound.

Let $P_\pm^{(n)}$ denote the adaptive transcript laws, including any
policy randomness.  The policy uses the same conditional kernel under
both alternatives.  The Gaussian conditional divergence and the KL
chain rule therefore give
\[
 D(P_+^{(n)}\|P_-^{(n)})
 =\sum_{j=1}^n\E_+
        \frac{(L_+(s_j)-L_-(s_j))^2}{2\sigma^2}
 \le \frac{nK_{\rm up}^2\delta^{2r}}{2\sigma^2}.
\]
The classical two-point testing inequality
$P_+(\mathrm{error})+P_-(\mathrm{error})\ge\tfrac12e^{-D(P_+\|P_-)}$
\citep{ref45} proves \eqref{eq:inverse-noise-lower}.
For the upper bound, the sample mean of repeated measurements at zero
has variance $\sigma^2/n$.  Thresholding it halfway between the two
known means has each error at most
$\exp[-nc_0^2\delta^{2r}/(8\sigma^2)]$.
\end{proof}

The exponent is sharp for this two-point experiment, with constants
depending on the fixed teacher and cycle.  The theorem gives neither
arbitrary-graph reconstruction nor uniform guarantees as dimension
increases.  Rational independence is not needed for this two-point
test, although it can be imposed to place both candidates inside
Theorem~\ref{thm:inverse-global}.
Cycle-dependent statistical difficulty has a close precedent in
determinantal point process learning \citep{urschel2017}.  Here the
additional statement is uniform over all nonnegative learning times
and hence over adaptive noisy loss-query policies.

\paragraph{Failures and symmetries.}
If $T=I_2$, a $45^\circ$ rotation of $M_0=\diag(2,3)$ gives an initial
matrix with diagonal entries $2.5$ and off-diagonal entries of magnitude
$0.5$.  These states are not diagonal-sign conjugate, but their entire
loss curves agree by orthogonal invariance.  Distinct teacher entries
alone do not justify the noncollision proof either: for $(1,2,3)$,
$4(t_1+t_3)=8t_2$.  That resonance invalidates direct extraction of the
two coefficients; it does not by itself prove nonidentifiability.
A chordless four-cycle, in contrast, is fully covered by
Theorem~\ref{thm:inverse-global}: its signs first appear at degree four,
even when no triangle or nonzero star is available.

\section{Exact certificates and numerical checks}
\label{sec:numerical}
The computations address admissibility, full-weight realization,
the singular limit, and the conditioning of the inverse problem.
They are deterministic examples.  Numerical integration discrepancies
are tolerance diagnostics, not rigorous interval error bounds.
Exploratory choices and all declared cells are retained in the
accompanying protocol and source data.

\subsection{A larger admissible perturbation on a fixed finite support}
For $q=r=3$, $c=1$, $\ell=(1/2,-1,1/2)$ and unit edge magnitudes,
we solve the finite moment problem on
$\{0,\pm\sqrt3\}^8$.  The desired fourth tensor is Gaussian except
\[
 \E[x_a^2x_jx_k]=\varepsilon\ell_a(B_s)_{jk},
 \quad a\in A,\quad j,k\in C,\ j\ne k.
\]
Auxiliary sign averaging and a common interaction sign flip reduce
the problem to 976 orbit columns and 140 constraints.
A linear program finds $\varepsilon_*=9/26$.
An exact rational basic solution has 111 positive orbit weights.
Its mass and moment residuals are exactly zero.  A separate rational
dual certificate has nonnegative slack on all 976 columns and value
$9/26$, proving the optimum on this specified support and symmetry
class.  We make no claim about the full moment cone.

Mixing this law with $1/5$ of the Gaussian-moment three-node product
law yields $\varepsilon=18/65$ and positive mass at all 6,561 grid
points.  The second arm is the branch-permutation image.
Independent expansion checks all 8 first, 36 second and 330 distinct
fourth monomials in each arm with exact rational residual zero.
Contracting the raw centered features gives the complete $36$-dimensional
operator; its maximum entrywise discrepancy from the analytic operator
is $3.09\times10^{-14}$, and the paired eigenvalue discrepancy is
$8.88\times10^{-16}$.  The explicit isometry fixes every diagonal
projector and has a nonzero Jordan defect of norm $\sqrt2$ in a
two-edge example.

\subsection{A finite-time contrast in the original weight dynamics}
\label{sec:numerical-weight}
The first tested teacher $(2,3,\ldots,9)$ produces a small maximum
common-loss gap, approximately $3.99\times10^{-7}$.
We retain it as evidence that a legal larger perturbation alone does
not guarantee an appreciable contrast.  A subsequent declared
108-cell teacher grid varies auxiliary offset and spacing, and
interaction offset and spacing.  The displayed selected cell is
\[
 T=\diag(1.1,31.1,61.1,2,5,8,11,14).
\]
It lies on the grid's largest auxiliary-spacing boundary and is
exploratory.  No global optimization or representative-task claim is made.

The selected cell has initial loss $2426.515$.
The maximal recorded common-loss gap is $0.178528$ near time $0.048398$,
or $7.36\times10^{-5}$ of initial loss.
Among recorded times where the mean current loss is at least
$10^{-4}$ of initial loss, the largest relative common-loss gap is
$4.351\%$ near time $0.196214$, at mean loss $0.256304$.
These two normalizations answer different questions; the larger
percentage does not describe the initial-loss scale.

Independent replay integrates the full $8\times8$ weight equation
using the raw finite-law operator, rather than the generating block
equation.  Tightening DOP853 tolerances changes the evaluated losses
by at most $2.27\times10^{-10}$, and $WW^\top$ agrees with a separately
integrated complete predictor equation to $6.04\times10^{-11}$.
At the fixed times $0.05,0.1,0.2$, the common-loss gaps are respectively
$0.176956,-0.0922444,0.0108301$.  Their signs change: there is no
uniform winner.  Actual weight gradient descent with step sizes
$10^{-4}$ and $5\times10^{-5}$ reproduces the gaps, with maximal
loss discrepancy from the flow decreasing from $9.85\times10^{-3}$
to $4.90\times10^{-3}$.

\begin{figure}[tb]
\centering
\includegraphics[width=\textwidth]{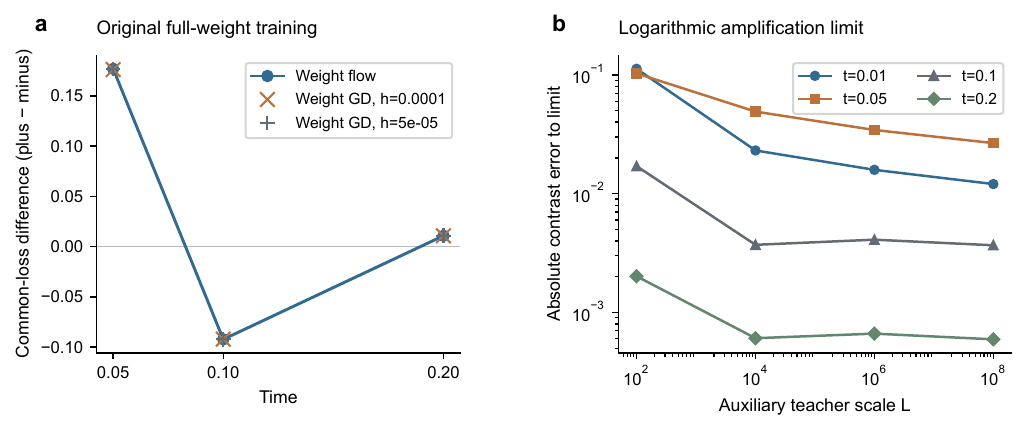}
\caption{Spectral equivalence, finite training contrast and its scale
dependence. \textbf{a}, Paired common-loss differences from complete
weight flow and actual weight updates at the three fixed observation
times.  Lines joining the flow observations only guide the eye.
\textbf{b}, Absolute errors between the common-loss contrasts
of the original coupled dynamics and the derived limiting contrasts
at four fixed times, as the auxiliary teacher scale grows.
Some errors are nonmonotone.  All four declared scales
are shown.  Both panels are deterministic checks without sampling
error bars.  The teacher in panel a was selected by the disclosed
108-cell exploration; panel b uses diverging teacher norms.}
\label{fig:geometry}
\end{figure}

\subsection{Analytic laws and scalar interventions}
For the trigonometric density in Proposition~\ref{prop:geo-smooth},
independent one-dimensional Gaussian quadrature followed by product
expansion checks all 495 monomials of degree zero through four in
each arm.  The largest target residual is $6.66\times10^{-16}$.
Characteristic-function derivatives provide an independent reference;
the largest reported quadrature error is $7.75\times10^{-12}$,
which is retained as the numerical resolution diagnostic.
With the unit triangle pair, $R_\psi=6e^2$ and
$\varepsilon=1/(12e^2)\approx0.011278$ certify a density ratio
between $1/2$ and $3/2$.  These analytic laws differ from the
larger-coupling discrete laws used for Figure~\ref{fig:geometry}a.

The scalar intervention is separately checked by raw-feature
contraction of the finite positive GH3 laws.
We use $\rho=0.2$, teacher $(2,3,\ldots,9)$ and couplings
$(18/65)/16,(18/65)/8,(18/65)/4,(18/65)/2$.
Every coupling includes the maximizing direction, a zero-intervention
control and three seeded bounded directions.
The largest discrepancy from the complete scalar signal formula is
$7.11\times10^{-15}$.
The same-input score in \eqref{eq:geo-paired-score} is evaluated
directly on every atom; its mean agrees with the normalized projected
gradient increment to $2.78\times10^{-17}$.
Its second moment under the GH3 product baseline is $0.04=\rho^2$.
These calculations check the score algebra and moments; the
finite-input error guarantee follows from the proof, not from a
Monte Carlo power estimate.

\subsection{The logarithmic-scale limit}
Set $T_A=(1.1,1.1+L,1.1+2L)$,
$T_C=\diag(2,5,8,11,14)$ and $\varepsilon=1/\log L$.
The four declared scales are $10^2,10^4,10^6,10^8$.
All couplings lie below the certified finite-support bound $9/26$.
Radau integration of the original coupled equations uses logarithmic
auxiliary coordinates and an analytic Jacobian.
An initial run using a numerical Jacobian completed but emitted
step-adjustment overflow warnings at the largest scale; the reported
rerun supplies the analytic Jacobian and treats runtime warnings as
errors.  The exact exposure identity is satisfied to at worst
$3.37\times10^{-11}$ across the checked states.

At time $0.05$, the common-loss gap rises from $0.164736$ at $L=10^2$
to $0.240929$ at $L=10^8$, compared with limiting gap $0.267555$.
The convergence is slow, as the logarithmic scaling suggests.
The other declared times include sign reversals and nonmonotone
errors (Figure~\ref{fig:geometry}b).  These checks support the
specified asymptotic formula.  The finite-$L$ error rate follows from
the proof of Theorem~\ref{thm:geo-amplification}, not from a fitted
numerical slope; neither result asserts amplification for bounded
teachers.  Directional finite-difference checks of the analytic
Jacobian are asserted only at $L=10^2$.  Their larger-$L$ values are
retained as unasserted cancellation-sensitive diagnostics.

\subsection{Finite inverse observations and their conditioning}
The inverse check uses dimension three, teacher
$(1,\sqrt7,\sqrt{73})$, six geometrically spaced observation times
from $0.003$ to $1$, and
\[
 C=0.02
 \begin{pmatrix}1&0.4&0.3\\0.4&-0.5&0.2\\0.3&0.2&0.7\end{pmatrix}.
\]
The exact observation Jacobian has smallest singular value
$2.45\times10^{-6}$ and condition number $2.22\times10^7$.
A directional finite-difference check agrees with its analytic
derivative to relative error $1.77\times10^{-10}$.
Eleven exact symbolic checks cover the diagonal, magnitude, triangle
and chordless-square coefficients.  Sign-gauge and repeated-teacher
rotation controls agree to numerical precision.

\begin{figure}[tb]
\centering
\includegraphics[width=\textwidth]{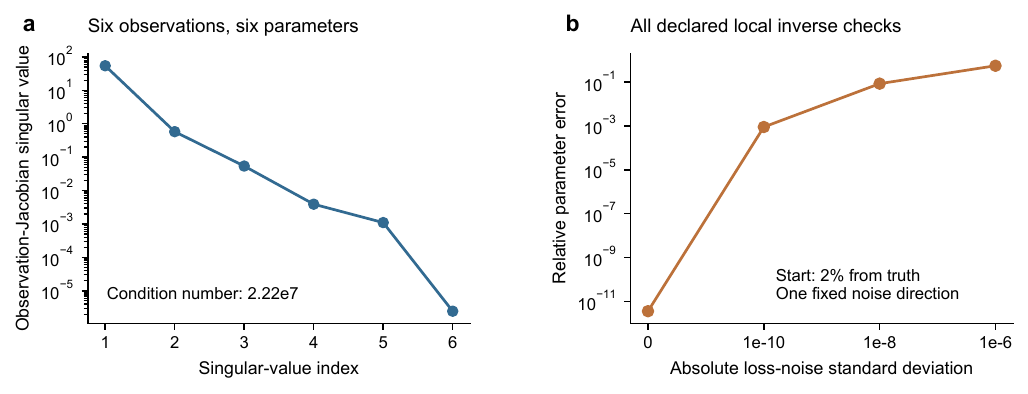}
\caption{Finite identifiability does not ensure stable reconstruction.
\textbf{a}, All six singular values of the analytic known-teacher
observation Jacobian in the specified three-dimensional example.
\textbf{b}, Relative parameter errors for every declared local inverse
attempt, including the noise-free control.  Each noise level uses the
same seeded Gaussian direction; these are single deterministic
realizations, not averaged performance estimates.  Initialization is
within $2\%$ of the true parameter in every case.}
\label{fig:inverse}
\end{figure}

All four locally initialized reconstruction attempts are reported in
Figure~\ref{fig:inverse}.  Starting $2\%$ from the true parameter,
the noise-free fit reaches relative parameter error
$3.62\times10^{-12}$.  Absolute observation-noise standard deviations
$10^{-10},10^{-8},10^{-6}$ give errors approximately
$8.96\times10^{-4},0.0846,0.553$, using one fixed noise direction.
The last noisy fit has a nonzero residual despite optimizer termination.
No ensemble accuracy, global convergence or noise-robust recovery
claim follows from these four cases.  Their purpose is to expose
the gap between an invertible observation map and a well-conditioned one.

For the blind extension we use the same teacher, replace $C_{22}/0.02$
by $0.8$ (one-based indexing), and fix nine geometric times from
$0.003$ to $1$.  The joint Jacobian contains six state and three teacher
columns.  Its determinant at 90-digit working precision is approximately
$-7.24704\times10^{-26}$; its smallest singular value is
$1.33\times10^{-7}$ and condition number $4.81\times10^8$.
Central differences agree with the analytic joint derivative to
relative error $3.71\times10^{-10}$, and independent high-precision
differentiation agrees to working precision.
An exact two-dimensional symbolic check gives zero residual in
the identity $L=Q/P^2$ used for the finite global bound.
Sign, permutation and simultaneous-rotation controls preserve the loss;
zero error and an uncoupled equilibrium coordinate demonstrate
the stated failures.  These are fixed algebraic and rank checks,
not a blind parameter-fitting experiment.

\section{Discussion}
The results distinguish three information regimes.  Before training,
even complete spectral data anchored on the whole diagonal teacher
algebra can miss multiplication relations that control nonlinear
evolution.  During training, these relations can remain weak to
arbitrarily high perturbative order, or acquire a finite effect when
a specified teacher scale diverges.  Bounded diagonal initialization
interventions expose the same hidden distinction at first order in
the coupling through one scalar gradient direction.  Shared-input
gradient differences implement pair testing without the teacher or
population-gradient access, with a sharp input-sample exponent.
After training, an exact scalar
history can encode the full hidden state modulo a natural symmetry.
The observation channel and parameter regime determine which
description is sufficient.

In the amplification limit, both laws
approach the same base law as their moment perturbations vanish, but
their teacher norms grow without bound.  On bounded teacher sets,
ordinary continuous dependence forces the trajectory contrast to
vanish.  The result identifies the nonuniform limit and its effective
initial state for the specified diverging-teacher family.
The numerical examples likewise retain their initial-loss
normalization and exploratory selection.

The inverse results separate exact information from usable estimates.
With a known nonresonant teacher, identifiability survives arbitrary
support sparsity.  On the stated active class the scalar curve also
identifies an unknown teacher spectrum, by recovering the generators
of its exponent monoid.  With distinct unordered teacher sums, the
joint finite-observation theorem gives a sharp generic local parameter
count in fixed dimension.  For a known teacher or the active blind
class with a dimension cap, the global bounds replace a continuum by
a finite exact design.  Noise amplification remains uncontrolled by
these counts.  Cycle information can be exponentially weak,
and the joint numerical Jacobian is poorly conditioned.
Stable and computationally feasible recovery under finite-sample loss
estimation, with unknown teachers or constrained graph classes,
remains open.

For neural learning theory, the constructive lesson is that the
linear operator and its task-anchored spectral measures are not the
whole dynamical object.  Matrix multiplication supplies additional
relations among responses.  The same relations that obstruct a
spectral forecast can become visible in a nonlinear trajectory.
Thus a nonlinear scalar trajectory can reveal multiplication-dependent
invariants absent from the matched spectral summaries.

\paragraph{Connection to information-organization sculpting.}
The sculpting perspective separates three questions: where information
is represented, which relations connect those representations, and
which combinations the network uses.  In the present model, the
additional relation is the Jordan product of the predictor matrix and
its loss gradient in the closed population dynamics.  The paired
construction matches the complete spectral measures anchored on the
diagonal teacher algebra while leaving multiplication-dependent cycle
information undetermined.  Both training trajectories remain in the
invariant predictor-state space described in Section~\ref{sec:geometry}.
The omitted relations have a demonstrable effect on the nonlinear
loss trajectory.

The broader hypothesis also concerns how learning reshapes feature
distributions and functional combinations, and how the resulting
organization affects later learning.  In our forward comparison,
the controlled distinction is introduced through the input laws;
architecture, teacher and full initialization are common to each
pair.  These results do not establish training-induced reorganization
of feature circuits or functional feature use in deep networks.
A direct test would branch from a common training checkpoint,
perturb feature relations while controlling feature strengths and
initial task performance, and predict both a change in the network's
dependence on those features and its subsequent trajectory.
Such tests would address the sculpting hypothesis beyond the
sufficiency result proved here.

\paragraph{Data and code availability.}
Versioned code and synthetic data accompany this manuscript as arXiv
ancillary files.  They include exact finite-law weights, rational
primal and dual certificates, every exploratory grid cell, full-weight
replay, the declared protocols, plotting scripts and a SHA-256 file
manifest.  Code is licensed under MIT and covered synthetic outputs
under CC BY 4.0.  No repository DOI has been assigned.

\paragraph{Funding and competing interests.}
The author reports no funding and no competing interests.

\paragraph{Use of AI assistance.}
AI tools assisted with theoretical exploration, drafting, coding,
literature discovery and independent computational checks.
The mathematical claims, computational results and references remain
the author's responsibility.

\appendix
\section{Initial-time order and discrete-step coefficients}
\label{app:local-order}
For completeness, this appendix supplies the coefficient calculation
used to show that the finite-time response in
Theorem~\ref{thm:geo-response} is nonzero.  Here the physical coupling
is denoted by $\beta=\varepsilon$ and the auxiliary barycentric vector
has the normalization in Section~\ref{sec:model}.
\subsection{The first teacher-weighted closed walk}
Put $D=\diag(g_C)$ and
\begin{equation}
 J_r=\tr\{D(B_+^r-B_-^r)\}
     =4\Pi_w\left(\sum_{i\in C_1\setminus C_2}g_i
                    -\sum_{i\in C_2\setminus C_1}g_i\right),
 \qquad \Pi_w=\prod_{j=1}^rw_j.
 \label{eq:isospectralJ}
\end{equation}
Choose $g$ so that $J_r\ne0$; increasing distinct negative integer magnitudes
along the two branches is one explicit choice. All traces of pure powers of
$B$ agree, but this joint invariant of the teacher and interaction does not.
The shared hub's two once-around contributions cancel in the difference.
Every non-hub cycle vertex gives $+4\Pi_w$ or $-4\Pi_w$. Shorter closed walks
have even parity on every edge and contribute no difference, including
walks with arbitrary diagonal stays.

\begin{lemma}[Series degree and closed-walk filtration]
\label{lem:isospectralfiltration}
Let the two laws, teacher and initialization be as above, and put
$\lambda=(r-1)q+1$. Use the original continuous block equation or, for
any fixed finite integer step count, the exact weight-gradient update
in Equation~\eqref{eq:isospectralGDupdate}. In the corresponding formal
variable $u=t$ or $u=h$,
\[
 \operatorname{val}_u(Q_+-Q_-)\ge\lambda,\qquad
 \operatorname{val}_u(a_+-a_-)\ge\lambda+1,
 \qquad
 \operatorname{val}_u(L_+^\zeta-L_-^\zeta)\ge rq,
 \quad \zeta\in\{G,P\}.
\]
Here a zero series has infinite valuation. At degree $\lambda$ in $Q$,
a differing closed matrix-spine walk must contain exactly $r$ edges and one
interaction-teacher insertion $D$. This assertion concerns the differing
maximal-spine coefficient: common scalar trace factors and common feedback
may contribute to $Q$, but do not change that coefficient.
\end{lemma}

\begin{proof}
Write $S_0=\ell^\top\rho$, where each Gaussian reference coordinate solves
\[
 \dot\rho_i=-4\rho_i(\rho_i-c+g_i),\qquad \rho_i(0)=c.
\]
Its coefficient $\rho_{i,m}$ is a polynomial in $g_i$ of degree at most
$m$, with leading term $c(-4)^mg_i^m/m!$. Thus
\[
 S_0=s_0t^{q-1}+O(t^q),\qquad
 s_0=\frac{c(-4)^{q-1}}{(q-1)!},\qquad
 \gamma=-\frac{2\beta cs_0}{q}.
\]
For the matrix filtration, first hold $S$ fixed as a common scalar series
of valuation at least $q-1$, mark every matrix-spine occurrence of $B$ by
$z$, and mark every spine occurrence of $D=\diag(g_C)$ by $\eta$. The
marked continuous recurrence is
\[
 \dot N=-4N^2+4cN-2\eta\{D,N\}-\beta S\{zB,N\}.
\]
If $N=\sum_{n\ge0}N[n]t^n$, its coefficient recurrence is
\[
 (n+1)N[n+1]=-4\sum_{a+b=n}N[a]N[b]+4cN[n]
 -2\eta\{D,N[n]\}-\beta\sum_{p+b=n}S[p]\{zB,N[b]\}.
\]
Induction in $n$ gives
\[
 \operatorname{val}_t[z^j\eta^d]N\ge jq+d.
\]
At the first possible degree, the $d=0$ coefficient is the pure word
\[
 [t^{jq}z^j]N=p_jB^j,\qquad
 p_j=\frac{c(\gamma/c)^j}{j!}.
\]
Indeed, the leading $S_0$ term supplies one $B$ and $q$ time degrees after
integration; all quadratic terms and all higher coefficients of $S$ occur
later. At degree $jq+1$, the $D$-free terms are still common scalar
multiples of $B^j$, while every non-pure term at this degree has exactly
one $D$ insertion.

Put
\[
 R_j=[t^{jq+1}z^j\eta]N.
\]
Keeping exactly one $D$ in the displayed recurrence yields
\[
 (jq+1)R_j=-2p_j\{D,B^j\}-\beta s_0\{B,R_{j-1}\},
 \qquad R_0=-4cD.
\]
For every $k\ge0$, cyclicity of trace and the relation defining $\gamma$
give
\[
 \tr(B^kR_j)=f_jc(\gamma/c)^j\tr(DB^{j+k}),\qquad
 f_0=-4,\qquad
 f_j=\frac{-4/j!+qf_{j-1}}{jq+1}=-\frac4{j!}.
\]
This is the insertion calculation, rather than a word-counting claim. In
particular, the only one-$D$ maximal-spine contribution to the first
differing scalar coefficient is obtained from $j=r-1$, $k=1$:
\[
 Q_+-Q_-=Vt^\lambda+O(t^{\lambda+1}),\qquad
 V=-\frac{4\gamma^{r-1}J_r}{c^{r-2}(r-1)!}.
\]

We now close the scalar feedback induction. If all earlier coefficients of
$Q_+$ and $Q_-$ agree, the scalar auxiliary recurrence makes every
coefficient of $S_+$ and $S_-$ needed for the current coefficient of $N$
agree as well. More explicitly, a differing $Q$ coefficient first enters
$a$ after one integration, and can first return through $S$ to $N$, hence
to $Q$, after a second integration. Therefore a coefficient of $Q$ at
degree $m$ can use a returned differing scalar only from a $Q$ coefficient
of degree at most $m-2$. Induction on $m$ consequently makes the feedback
common through degree $\lambda$, and gives
$a_+-a_-=O(t^{\lambda+1})$.

This induction does not say that common feedback is zero. For clarity,
the first common terms are
\[
 [t^q]Q=\gamma\tr(B^2),\qquad
 [t^{q+1}](S-S_0)=-\frac{2\beta c\gamma}{q+1}\|\ell\|^2\tr(B^2),
\]
and this correction induces the common contribution
\[
 [t^{q+2}]Q\big|_{S-S_0}
 =\frac{4\beta^2c^2\gamma}{(q+1)(q+2)}\|\ell\|^2
   \bigl(\tr(B^2)\bigr)^2.
\]
It is generally nonzero, but it is identical for the two arms and hence
cancels in their difference. Thus common feedback is excluded only from
the differing maximal-spine coefficient above, not from $Q$ itself.

It remains to identify which common coefficient can differ. On the
two-cycle graph, the edges of a closed matrix-spine word that are traversed
an odd number of times form a cycle-space element. Fewer than $r$ closed
spine edges are therefore switching-even. With exactly $r$ edges and no
$D$, the contribution is a pure trace and agrees because $B_-$ is
conjugate to $B_+$. The outer $B$ in $Q=\tr(BN)$ is free, so the maximal
spine at degree $\lambda=(r-1)q+1$ has $j=r-1$ state edges. The preceding
recurrence shows that its sole differing possibility is exactly one $D$,
which gives the displayed formula for $V$. An already differing $Q$ cannot
alter this conclusion by the two-degree return just proved.

For the loss bounds, the interaction part of the common Gaussian risk has
at degree $rq$ only the pairing of $D$ with the pure-state $j=r$ term. It
equals $Z/r$, where
\[
 Z=\frac{\gamma^rJ_r}{c^{r-1}(r-1)!}.
\]
The auxiliary linear response around $\rho$ is
\[
 \Delta L_A=-2\beta\int_0^t\Delta Q(u)\mathcal K(t,u)\,du
 +\text{terms of later degree},\qquad
 \mathcal K(t,u)=\sum_{i\in A}\ell_i(\rho_i(t)-T_i)
 \Phi_i(t,u)\rho_i(u),
\]
where $\Phi_i$ is the scalar reference variational propagator.
To derive the relevant node-polynomial degree, put
\[
 b_i(v)=1+\frac c{T_i}(e^{4T_i v}-1),\qquad T_i=c-g_i.
\]
The exact scalar solution and propagator are
\[
 \rho_i(v)=\frac{ce^{4T_i v}}{b_i(v)},\qquad
 \Phi_i(t,u)=e^{4T_i(t-u)}\frac{b_i(u)^2}{b_i(t)^2}.
\]
Since $\rho_i(t)-T_i=g_i/b_i(t)$, the individual kernel summand is
\[
 (\rho_i(t)-T_i)\Phi_i(t,u)\rho_i(u)
     =cg_i e^{4(c-g_i)t}\frac{b_i(u)}{b_i(t)^3}.
\]
The coefficient of $v^n$ in $b_i(v)-1$ is
$c4^n(c-g_i)^{n-1}/n!$ for $n\ge1$, so its degree in $g_i$ is
at most $n-1$.  At bivariate total degree $m$, every term involving
a nonconstant coefficient of $b_i(u)/b_i(t)^3$ has node degree at
most $m$.  The sole node-degree-$m+1$ part is therefore
\[
 \frac{c(-4)^m}{m!}g_i^{m+1}t^m.
\]
Thus the weighted sum vanishes for $m<q-2$.  At $m=q-2$,
$\sum_i\ell_i g_i^{q-1}=1$ retains the displayed highest-degree term
and annihilates all the others, giving
\[
 \mathcal K(t,u)=\frac{c(-4)^{q-2}}{(q-2)!}t^{q-2}
 +O_{\rm total}(q-1).
\]
Substitution of the displayed formula for $V$ gives the auxiliary
contribution
\[
 Z\,\frac{q(q-1)}{(r-1)q+2}\,t^{rq}+O(t^{rq+1}).
\]
Common feedback changes this kernel only above the displayed degree, and
nonlinear arm differences have valuation at least $2\lambda+2$. Finally,
the own-law correction has
\[
 \frac\beta2\Delta(SQ)=qZt^{rq}+O(t^{rq+1});
\]
all its other difference terms start later. These facts prove the
continuous loss valuations in the statement.

For completeness, the finite-step bookkeeping is also coefficientwise.
At step $K$, set
\[
 G_K=N_K-cI+\eta D+\frac\beta2S_KzB,\qquad
 N_{K+1}=N_K-2h(G_KN_K+N_KG_K)+4h^2G_KN_KG_K.
\]
Thus $[h^nz^j\eta^d]N_{K+1}$ is its previous value plus $-2$ times the
two-factor convolution at $h^{n-1}z^j\eta^d$, plus $4$ times the
three-factor convolution at $h^{n-2}z^j\eta^d$. The direct $D$ term has
$h$-degree zero and weight one; the $B$ forcing has $S_K$-degree at least
$q-1$ and weight $q$; the prefactors $h$ and $h^2$ supply the missing one
and two degrees. Induction over $n$ in these explicit convolutions gives
zero for $n<jq+d$, a step-count-dependent scalar multiple of $B^j$ at
the lowest degree $jq$, and the first possible one-$D$ term at $jq+1$.
These discrete coefficients need not equal the continuous coefficients
at a fixed step count. The scalar update
\[
 a_{i,K+1}=a_{i,K}-4ha_{i,K}E_{i,K}+4h^2a_{i,K}E_{i,K}^2,
 \qquad E_{i,K}=a_{i,K}-T_i+\tfrac\beta2\ell_iQ_K,
\]
gives the same two-step feedback induction. Its linearized kernel has the
same degree-at-most-$m+1$ property in $g_i$, so barycentric annihilation
and the preceding loss calculation give no discrete loss coefficient below
$rq$.

Write the exact map as $M_{K+1}=M_K+hF(M_K)+h^2J(M_K)$, and let $m_n(K)$
denote any coefficient at $h^n$. Directly from the map,
\[
 m_n(K+1)-m_n(K)=[h^{n-1}]F(M_K)+[h^{n-2}]J(M_K).
\]
The first term has degree at most $n-1$ in $K$ and the second at most
$n-2$. Hence $\deg_Km_n\le n$; its highest $K^n$ coefficient comes only
from $F$ and is the continuous Taylor coefficient. After composition with
either loss, the leading $K^{rq}$ coefficient is therefore the continuous
$C^\zeta_{q,r}$ displayed below. This is an algebraic coefficient
induction for all $q,r\ge3$, not an inference from numerical cells.
\end{proof}

\subsection{Loss coefficients and actual weight updates}
With $\gamma=(-4)^qc^2\beta/(2q!)$ and
$Z=\gamma^rJ_r/[c^{r-1}(r-1)!]$, the preceding calculation gives
\begin{align}
 L_+^\zeta(t)-L_-^\zeta(t)&=C^\zeta_{q,r}t^{rq}+O(t^{rq+1}),\nonumber\\
 C^G_{q,r}&=Z\left[\frac1r+\frac{q(q-1)}{(r-1)q+2}\right],\nonumber\\
 C^P_{q,r}&=Z\left[q+\frac1r+\frac{q(q-1)}{(r-1)q+2}\right].
 \label{eq:isospectralcoefficients}
\end{align}
Both coefficients are nonzero when $J_r\ne0$.
Analyticity of the polynomial vector field justifies the local expansions.
The exact weight-gradient map used in the proof is
\begin{equation}
 M_{s,k+1}=M_{s,k}-2h\{G_s(M_{s,k}),M_{s,k}\}
             +4h^2G_s(M_{s,k})M_{s,k}G_s(M_{s,k}).
 \label{eq:isospectralGDupdate}
\end{equation}
For fixed integer $k$, both loss differences have no $h$ coefficient
below degree $rq$.  The degree-$rq$ coefficient is a polynomial in $k$
of degree exactly $rq$, with leading coefficient $C^\zeta_{q,r}$.
It is therefore nonzero for all but finitely many positive integer
step counts.  This assertion concerns small step size at fixed step
count, not long-time or stochastic-training performance.

\bibliographystyle{plainnat}
\begingroup
\small
\setstretch{1.0}
\setlength{\bibsep}{3pt}
\bibliography{references}
\endgroup
\end{document}